\documentclass[a4paper,fleqn]{cas-dc}

\usepackage[numbers,compress,sort]{natbib}
\usepackage{supertabular}
\usepackage{amsmath}
\usepackage{bm}
\usepackage{float}
\usepackage{multirow}
\usepackage{graphicx}
\usepackage{makecell}
\usepackage[normalem]{ulem}
\useunder{\uline}{\ul}{}
\usepackage{xcolor}
\usepackage{subfigure}
\usepackage{subcaption}
\usepackage{booktabs,arydshln}
\usepackage{xurl}
\usepackage{enumitem}
\usepackage{algorithm}
\usepackage[noend]{algpseudocode}

\def\tsc#1{\csdef{#1}{\textsc{\lowercase{#1}}\xspace}}
\tsc{WGM}
\tsc{QE}
\tsc{EP}
\tsc{PMS}
\tsc{BEC}
\tsc{DE}

\newdefinition{mydef}{Definition}

\begin{document}
\let\WriteBookmarks\relax
\def\floatpagepagefraction{1}
\def\textpagefraction{.001}

\shorttitle{Enhancing Noise Robustness via Contrastive Feature Augmentation}

\shortauthors{Z. Tang et~al.}

\title [mode = title]{Enhancing Noise Robustness of Parkinson’s Disease Telemonitoring via Contrastive Feature Augmentation}



%

\author[1]{Ziming Tang}[style=chinese]
\ead{tzm24@mails.tsinghua.edu.cn}

\author[2]{Chengbin Hou}[style=chinese,orcid=0000-0001-6648-793X]
\cormark[1]
\ead{houcb@fyust.edu.cn}

\author[1]{Tianyu Zhang}[style=chinese]

\author[2]{Bangxu Tian}[style=chinese]


\author[3]{Jinbao Wang}[style=chinese,orcid=0000-0001-5916-8965]
\ead{wangjb@szu.edu.cn}


\author[1]{Hairong Lv}[style=chinese,orcid=0000-0003-1568-6861]
\cormark[1]
\ead{lvhairong@tsinghua.edu.cn}

\affiliation[1]{organization={Ministry of Education Key Laboratory of Bioinformatics, Department of Automation},
    organization={Tsinghua University},
    city={Beijing},
    postcode={100084}, 
    country={China}}
    
\affiliation[2]{organization={School of Computing and Artificial Intelligence},
    organization={Fuyao University of Science and Technology},
    city={Fujian},
    postcode={350109}, 
    country={China}}

\affiliation[3]{organization={National Engineering Laboratory for Big Data System Computing Technology},
    organization={Shenzhen University},
    city={Shenzhen},
    postcode={518060}, 
    country={China}}

\cortext[cor1]{Corresponding authors: Chengbin Hou, Hairong Lv}



\begin{abstract}
Parkinson’s disease (PD) is one of the most common neurodegenerative disorder. PD telemonitoring emerges as a novel assessment modality enabling self-administered at-home tests of Unified Parkinson’s Disease Rating Scale (UPDRS) scores, enhancing accessibility for PD patients. However, three types of noise would occur during measurements: (1) patient-induced measurement inaccuracies, (2) environmental noise, and (3) data packet loss during transmission, resulting in higher prediction errors. To address these challenges, NoRo, a noise-robust UPDRS prediction framework is proposed. First, the original speech features are grouped into ordered bins, based on the continuous values of a selected feature, to construct contrastive pairs. Second, the contrastive pairs are employed to train a multilayer perceptron encoder for generating noise-robust features. Finally, these features are concatenated with the original features as the augmented features, which are then fed into the UPDRS prediction models. Notably, we further introduces a novel evaluation approach with customizable noise injection module, and extensive experiments show that NoRo can successfully enhance the noise robustness of UPDRS prediction across various downstream prediction models under different noisy environments. 
\end{abstract}




\maketitle

\section{Introduction}\label{sec:intro}

Parkinson's Disease (PD) is the second most common age-related neurodegenerative disorder after Alzheimer's disease \cite{kalia2015parkinson}. A combination of aging, genetic predipositions, and environmental factors are known contributors to the development of PD \cite{ascherio2016epidemiology,bloem2021parkinson}. Among these, aging is the most significant risk factor for PD. Therefore, as the global population ages, the prevalence of PD is expected to rise steadily, exacerbating societal health and economic challenges \cite{reeve2014ageing}. For instance, by 2004, the prevalence of PD had surpassed 1\% among individuals over 60 years old, and by 1998, more than 1 million people in North America were diagnosed with PD \cite{samii2004parkinson, ae1998parkinson}. The 41st Healthy China Huaxi Health Forum reported that by the end of 2021, China had nearly 3 million PD patients with 100,000 new cases annually.

Thus, monitoring the progression of PD has attracted the attention worldwide. Various methods based on the pathological characteristics (e.g., the presence of abnormal Lewy bodies) have been proposed \cite{halliday2010progression}. In addition to these pathological characteristics, PD is associated with many clinical manifestations, where motor symptoms are considered the cardinal signs of PD \cite{jankovic2008parkinson}. Accordingly, clinical scales such as Unified Parkinson’s Disease Rating Scale (UPDRS) are also employed to capture clinical features and monitor the propagation of PD \cite{poewe2009clinical}. However, monitoring PD progression typically requires patients' to visit the hospital. For individuals with motor symptoms such as movement disorders and gait difficulties, frequent hospital visits can be both inconvenient and challenging. 
To address this, a non-invasive telemonitoring approach has been developed, enabling patients to assess PD progression at home. 

This approach utilizes Intel Corporation's At-Home Testing Device (AHTD) to capture speech data from PD patients, with the goal of measuring motor impairment symptoms associated with PD \cite{goetz2009testing}. After speech signal processing, 16 features are extracted from the patients' speech patterns, which are then mapped to UPDRS scores \cite{tsanas2009accurate}. UPDRS is a widely recognized and validated clinical rating scale for PD, extensively used to assess disease progression and providing comprehensive coverage of motor symptoms \cite{movement2003unified,nilashi2018hybrid}. 
There have been several works utilize these speech features to predict UPDRS scores. Most works employ hybrid architectures for the prediction task. Hybrid systems composed of clustering methods such as Self-Organizing Maps (SOM) and Expectation-Maximization (EM), along with regression methods such as Gaussian Process Regression (GPR) and Adaptive Network-based Fuzzy Inference System (ANFIS), have been proposed \cite{nilashi2019analytical,nilashi2023parkinson, zogaan2024combined, vats2023predicting}.

However, as PD patients use the AHTD to conduct speech tests at home without professional supervision, various sources of noise would affect the accuracy of testing results. First, the AHTD requires PD patients to maintain a distance of approximately 5 centimeters from the microphone and produce vowels at a consistent frequency, which is challenging for elder people to achieve consistently \cite{goetz2009testing}. Second, environmental noise may interfere with the clarity of the speech recordings, further compromising results \cite{nicastri2004multidimensional}. Third, during data processing, the collected speech data must be encrypted and transmitted to a server for analysis using speech signal processing algorithms. Issues such as packet loss or decryption can occur during data transmission or decryption, potentially affecting the reliability of the results \cite{goetz2009testing}. Noise increases the randomness and instability of predictions, leading to predictions that deviate from the true UPDRS scores. Previous studies have generally overlooked these noisy scenarios, despite achieving some success in UPDRS prediction tasks. 

To address these challenges, a \underline{No}ise-\underline{Ro}bust (namely NoRo) UPDRS prediction framework is proposed. First, speech features are grouped into some ordered bins based on the continuous values of a feature selected by a feature selection algorithm. Second, Contrastive Learning (CL) is applied to generate noise-robust features. Specifically, by treating the same-bin features as positive pairs and cross-bin features as negative pairs, CL is employed to train a Multilayer Perceptron (MLP) encoder to project original features as hidden states. Finally, the noise-robust features (i.e., hidden states) are concatenated with the original speech features as the augmented features, which are then fed into downstream regression models for predicting UPDRS scores. 

Intuitively, with NoRo framework, the samples (patients) with similar features in the original feature space get closer in the augmented feature space, whereas they are pushed away from each other if the similarity is low. As a result, the augmented features become more robust to potential noise, since the discriminative nature of these samples is preserved in the augmented feature space even under some noisy environments, thereby enhancing the performance and robustness of downstream machine learning tasks. To evaluate the effectiveness and robustness of NoRo, we further propose an evaluation approach with customizable noise injection module, and test various noisy environments with different UPDRS prediction models and settings using a real-world PD telemonitoring dataset. 
The main contributions of this work are summarized as follows.
\setlist{nolistsep}
\begin{itemize}[noitemsep]
    \item This work, for the first time, identifies the robustness issues in PD telemonitoring and discusses why measurement inaccuracies, environmental noise, and transmission loss may affect the UPDRS prediction.
    \item To address the robustness issue, a novel noise-robust UPDRS prediction framework (NoRo) is proposed. The idea is to divide continuous values into ordered bins such that the contrastive learning can be used to learn noise-robust features without human labeling. NoRo is a flexible framework and can be freely applied to various UPDRS prediction models.
    \item We further introduce a novel evaluation approach with customizable noise injection module. Extensive experiments are conducted to demonstrate the effectiveness and robustness of the proposed NoRo. It is worth noting that NoRo reduces the prediction errors by up to more than 10\%-40\% in noisy environments.
    \item To benefit future research, the source code is publicly available at https://github.com/tzm-tzm/PD-Robust.
\end{itemize}
\section{Related Work}
\subsection{PD Diagnosis}
aaaaaa

\subsection{PD Telemonitoring} 
bbbbbb

\subsection{Applications of Contrastive Learning(CL)}

\section{Methodology}\label{method}
\begin{figure*}[htbp]
	\centering
	\includegraphics[width=0.995\linewidth]{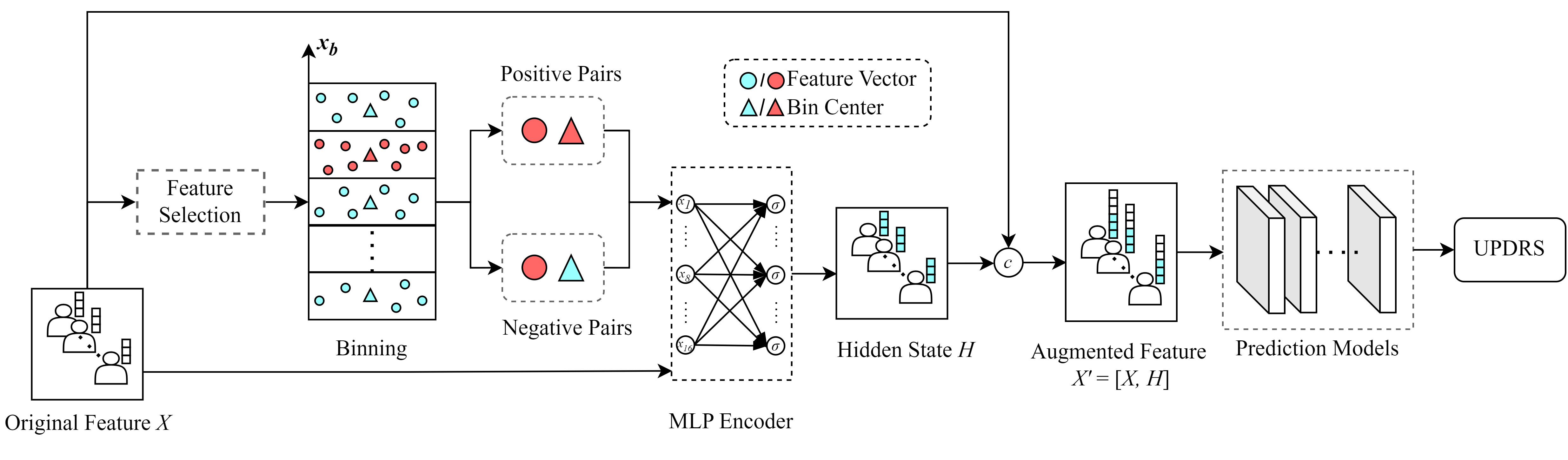}
	\caption{NoRo, a framework of the noise-robust UPDRS prediction process. NoRo enhances prediction robustness through a self-supervised Contrastive Learning (CL) approach that generates noise-robust augmented features. First, a Random Forest algorithm selects the feature dimension $\bm{x}_b$ with the highest importance score across the original speech feature $X$, maximizing correlation with UPDRS scores. Second, $\bm{x}_b$ undergoes equal-width binning to group $X$ into $K$ bins. Third, following the idea of CL, same-bin features are treated as positive pairs and cross-bin features as negative pairs to train a Multilayer Perceptron (MLP) encoder $W$ to project $X$ as hidden states $H=\sigma(WX)$. Then, $H$ is concatenated with $X$ as the augmented feature $X'=\left[X,H\right]$. Finally, augmented features are then fed into downstream prediction models to achieve robust UPDRS prediction.}
	\label{fig:overview}
\end{figure*}


\subsection{Problem Formulation}

Let the speech feature matrix be $X=[\bm{x}_1,\dots,\bm{x}_M]^T\in \mathbb{R}^{M\times D}$ with $M$ samples and $D$ dimensions per sample. $X$ with unknown inherent noise $N\in \mathbb{R}^{M\times D}$ can be modeled as $X=Z+N$, where $Z\in \mathbb{R}^{M\times D}$ is the true speech feature.

\begin{mydef}
	(Feature Augmentation): $X$ can be projected as hidden states (noise-robust features) $H=\sigma(XW)\in\mathbb{R}^{M\times D'}$ using an encoder $W$, where $D'$ is the dimension of the hidden states. Then, $H$ can be concatenated with $X$ as the augmented feature $X'=\left[X,H\right]\in\mathbb{R}^{M\times (D+D')}$. The process to generate $X'$ from $X$ is called \textit{Feature Augmentation}.
\end{mydef}

\begin{mydef}
	\label{Def:Robustness}
	(Noise Robustness): Noise robustness of a regression model refers to its ability to tolerate noise. It can be measured by the prediction error in noisy environment, where lower error indicates stronger noise robustness.
\end{mydef}

\begin{mydef}
    (PD Telemonitoring): PD telemonitoring is achieved by predicting both Motor and Total UPDRS scores through remotely collected speech features $X$ from PD patients. UPDRS score $\bm{y}$ is a continuous value. The prediction of $\bm{y}$ requires the regression model $P$ to utilize several continuous features $X$, yielding the predicted value through $\hat{\bm{y}}= P(X)$.
\end{mydef}

The goal of feature augmentation is to enhance the robustness of UPDRS prediction when speech features $X$ are contaminated by noise. Using the augmented features $X'$, lower prediction errors are achieved compared to using $X$ under identical noise conditions.





\subsection{Prediction Framework NoRo}

A prediction framework NoRo based on contrastive feature augmentation is proposed in this work to enhance downstream models' noise robustness. The augmented features become more robust to potential noise, thereby enhancing the prediction performance.

\subsubsection{Data Preprocessing}

The PD telemonitoring dataset consists of 16 speech features. However, the scales vary across the 16 features. For instance, feature \{HNR\} reaches a scale of $10^1$, while feature \{Jitter(Abs)\} reaches a scale of $10^{-5}$ to $10^{-6}$. Due to this significant variation in the orders of magnitude, a normalization process is necessary to eliminate this scale-induced bias prior to experimental analysis. In this work, a z-score normalization is employed as shown in Eq. (\ref{eq:zscore}).

\begin{equation}
    x_{ij} \leftarrow \frac{x_{ij}-\overline{x}_j}{\sigma_{x_j}},
    y_{i} \leftarrow \frac{y_{i}-\overline{y}}{\sigma_{y}}
    \label{eq:zscore}
\end{equation}
Here, $x_{ij}$ is the $j$-th feature of the $i$-th sample, $\overline{x}_j$ and $\sigma_{x_j}$ are the mean and the standard deviation of the $j$-th feature across all samples. The normalization of label $y$ is similar to that of $x$. The z-score normalization process is applied to both training and testing datasets, using the mean and standard deviation values of the training dataset to prevent data leakage.

\subsubsection{Feature Selection Module}
\label{sec:feature selection}

After the data preprocessing progress, a Random Forest (RF) algorithm is employed to select the binning feature $\bm{x}_b\in\mathbb{R}^{M\times1}$ by assessing the importance of each feature.

RF is an ensemble learning method that assesses feature importance by aggregating the contribution of each feature across all decision trees. RF has been utilized to select key features, improving the identification of computer security threats by guiding the initialization of the searching model \cite{hasan2016feature}.

In this work, RF is also employed to assess the importance score of each feature for both Motor and Total UPDRS according to Mean Decrease in Impurity (MDI). The feature that ranks highest for both UPDRS scores is chosen as the binning feature $\bm{x}_b$ following Eq. (\ref{eq:RF}). The results are detailed in Appendix \ref{sec:RF results}.

\begin{equation}
    \bm{x}_b = \arg\max\limits_{i\in\{1,\dots,D\}}(\text{MDI}(\bm{x}_i))
    \label{eq:RF}
\end{equation}

\subsubsection{Binning Module} \label{section:Binning}

Contrastive learning (CL) is employed in the feature augmentation method. To prepare positive and negative pairs for CL, equal-width binning is applied to $\bm{x}_b$. The range of $\bm{x}_b$ is divided into $K$ intervals of equal width. Feature vector $\bm{x}_i^T\in\mathbb{R}^{D\times1}$, whose $x_{ib}$ falls within the $k$-th interval is assigned to the $k$-th ($k\in\{1,\cdots,K\}$) bin, as illustrated in Eq. (\ref{eq:binning}). 

\begin{equation}
bin(\bm{x}_i) =
\begin{cases}
    k & \text{if } x_{ib} \geq \min{(\bm{x}_b)}+(k-1)\cdot\frac{\max{(\bm{x}_b)}-\min{(\bm{x}_b)}}{K} \\
    &\text{ and } x_{ib} < \min{(\bm{x}_b)}+k\cdot\frac{\max{(\bm{x}_b)}-\min{(\bm{x}_b)}}{K}, \\
    K& \text{otherwise.}
\end{cases}
\label{eq:binning}
\end{equation}




\subsubsection{Contrastive Learning Module}

To project $X$ to $H$, a Multilayer Perceptron (MLP) $W$ is trained as an encoder through CL. Hidden states $H$ can be obtained by $H=\sigma(XW)$, where $W\in\mathbb{R}^{D\times D'}$ is the projection matrix and $\sigma$ is Hyperbolic Tangent (Tanh) activation function. 

In this work, $D'$ is set equal to $D$. Thus, the augmented feature can be represented as $X'=[X,H]\in\mathbb{R}^{M\times 2D}$, and we have $H\in\mathbb{R}^{M\times D}$, $W\in\mathbb{R}^{D\times D}$.

\paragraph{\textbf{Contrastive Loss Function.}}

To train the MLP encoder, CL is employed to bring feature vectors within the same bin closer together, while pushing feature vectors from different bins farther apart in the projected feature space.

The loss function used for CL is the contrastive loss, as shown in Eq. (\ref{eq:CLLoss}):

\begin{equation}
	L = -\sum_{i=1}^{K}\sum_{j\in bin_i} log{\frac{\exp{(\bm{h}_j^T \bm{c}_i)}}{\sum_{k=1}^{K} \exp{(\alpha_{ik}\bm{h}_j^T\bm{c}_k)}}}
	\label{eq:CLLoss}
\end{equation}
Here, $\bm{h}_j\in\mathbb{R}^{D\times1}$ represents the $j$-th feature vector of $H$, and $\bm{c}_i=\frac{1}{M_i}\sum_{j\in bin_i}\bm{h}_j\in\mathbb{R}^{D\times1}$ represents the center of the $i$-th bin, which is defined as the mean of $\bm{h}$ within this bin.

By using this loss function, the similarities of the feature vectors within the same bin will be maximized and the similarities of the cross-bin samples will be minimized.


\paragraph{\textbf{Calculation of Bin Centers.}}

For the bins that contain at least one feature vector, the bin centers can be calculated by the mean value of the $\bm{h}$ belonging to them. However, when $K$ is more than 20, some bins may not contain any feature vectors, making it impossible to calculate the bin center using the previous method. 

For these bins that do not contain any feature vector, the bin center will be replaced by the bin center of the nearest non-empty bin. If there are two nearest non-empty bins, the bin center is represented by the average of their bin centers. Thus, bin center $\bm{c}_i$ of the $i$-th bin can be calculated by Eq. (\ref{eq:bin center}).

\begin{equation}
    \bm{c}_i=
    \begin{cases}
        \frac{1}{M_i}\sum\limits_{j\in bin_i}\bm{h}_j & \text{if }M_i>0,\\
        \bm{c}_{i+k} & \text{else if }\{M_{i-k},\dots,M_{i+k-\frac{k}{|k|}}\}=0 \\
        &\text{, and }M_{i+k}>0\text{, and } k\neq0,\\
        \frac{(\bm{c}_{i+k}+\bm{c}_{i-k})}{2} & \text{otherwise.}\\
    \end{cases}
    \label{eq:bin center}
\end{equation}
Here, $M_i$ indicates the number of samples in the $i$-th bin.

\paragraph{\textbf{Distance Coefficient.}}

Considering that bins are closer to each other have a stronger correlation, a distance coefficient is introduced to represent this relationship. $\alpha_{ik}$ represents the distance coefficient between the $i$-th and the $k$-th bin. 

\begin{figure}
	\centering
	\includegraphics[width=0.995\linewidth]{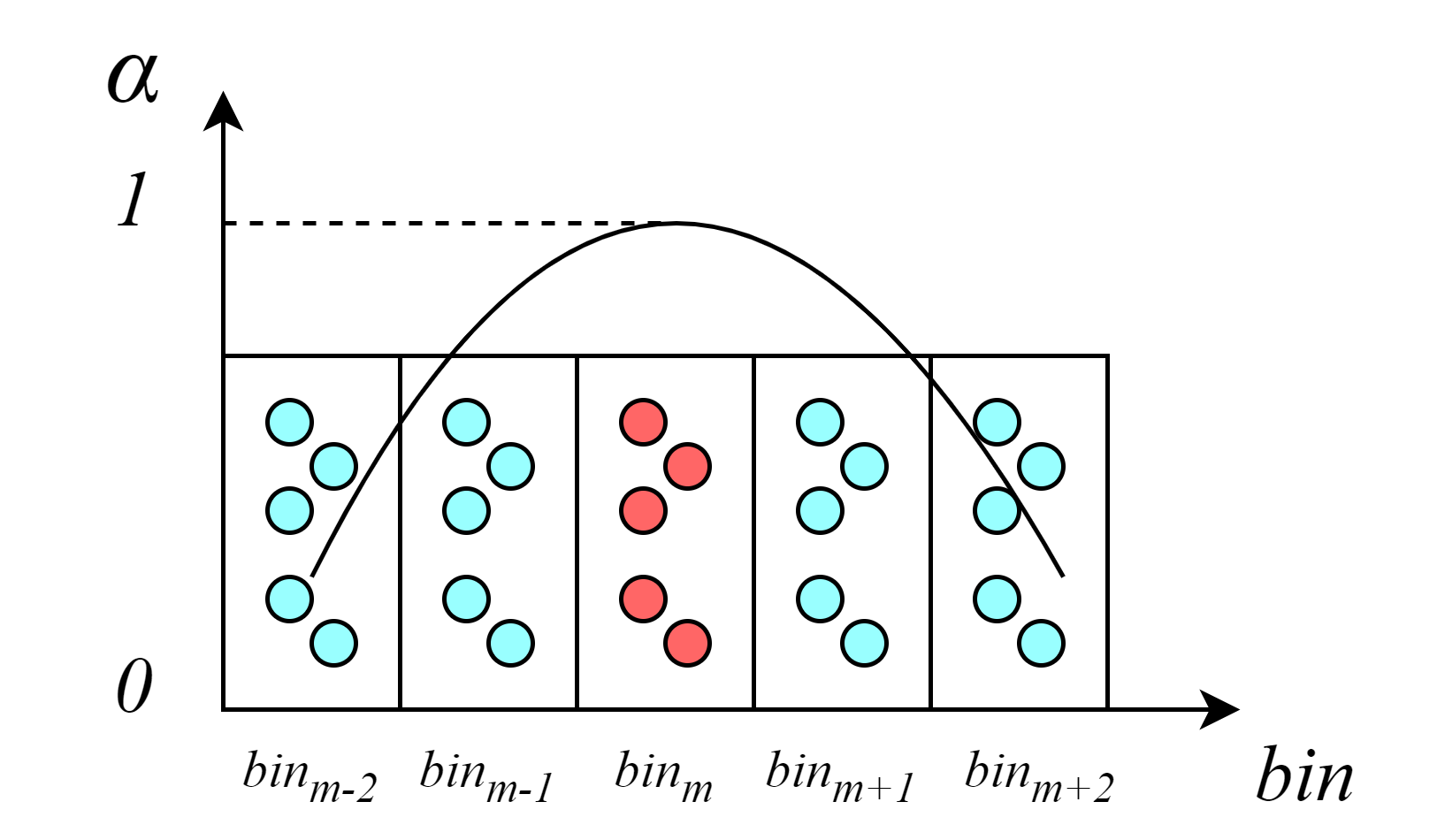}
	\caption{Distance Coefficient $\alpha$ for the $m$-th Bin. The curve represents the value of $\alpha$: (1) Same-bin $\alpha$ equals to $1$, which is the highest, e.g., $\alpha_{mm}=1$. (2) $\alpha$ decreases as the distance between the $m$-th bin and other bins increases, e.g., $0<\alpha_{m,m-2}<\alpha_{m,m-1}<\alpha_{m,m}$. (3) $\alpha$ is symmetric around the central bin, e.g., $\alpha_{m,m-1}=\alpha_{m-1,m}$ and $\alpha_{m,m-2}=\alpha_{m-2,m}$.}
	\label{fig:alpha-bin}
\end{figure}

The distribution of $\alpha$ obeys three rules: (1) Same-bin $\alpha$ equals $1$, which is the highest. (2) $\alpha$ decreases as the distance between bins increases but always > 0. (3) $\alpha$ is the same for all bins that are equidistant from the central bin, which needs to be symmetric around the central bin. $\alpha$ of the $m$-th bin is illustrated in Fig. \ref{fig:alpha-bin}.

For the design of $\alpha$, a normalized modified binomial distribution is applied in this work. $\alpha_{m,n}$ is shown in Eq. (\ref{eq:cal alpha}):

\begin{equation}
	\centering
	{\alpha}_{m,n}={\alpha}_{n,m}=\frac{\binom{N}{n-m+N/2}}{\binom{N}{N/2}}
	\label{eq:cal alpha}
\end{equation}
Here, $N$ is an even number to ensure the distribution function has a single maximum value, which occurs when $m=n$, expressed as $N=2\cdot max{(m, K-m)}$. Further normalization is applied to ensure the distribution reaches its maximum value of 1 when $m=n$. Due to the properties of binomial coefficients, this distribution is symmetric around $m$.

\subsection{Algorithm and Complexity}

The CL training algorithm of the MLP encoder is shown in Alg. \ref{alg:training}. Further implementation details are reported in Appendix \ref{app:MLP Training}.

\begin{algorithm}
	\caption{MLP Encoder Contrastive Learning Process}
	\label{alg:training}
	\small
	\begin{algorithmic}[1]
		\Require Speech Feature Matrix $X$; Bin Number $K$
		\Ensure Parameters of the Projection Matrix of MLP $W$
		
		\State Select the binning vector $\bm{x}_b$
		\State Perform binning based on $\bm{x}_b$
		\State Calculate distance coefficients $\{{\alpha}_{1,1},\ldots,{\alpha}_{K,K}\}$
		\State Random Initialize $W$
 		\For{$t$ = 1 to $T$}
                \State $H\leftarrow\sigma(XW)$
			 \For{$bin_i$ in $\{bin_1,bin_2,\ldots, bin_{K}\}$}
			 \State Calculate $\bm{c}_i$
			 \EndFor
			 \State Initialize loss $L=0$
			 \For{$\bm{h}_i$ in $H$}
			 	\State $L_i \leftarrow -log{\frac{\exp{(\bm{h}_i^T\bm{c}_i)}}{\sum_{k=1}^{K} \exp{(\alpha_{ik}\bm{h}_i^T\bm{c}_k)}}}$
			 	\State $L \leftarrow L+L_i$
			 \EndFor
			 \State $W\leftarrow W - \eta_{t}\cdot\nabla_W L(W)$
		\EndFor
		\Return $W$
	\end{algorithmic}
\end{algorithm}

The algorithm processes each sample to perform binning operations, compute hidden states and bin centers. This step exhibits a computational complexity of $\mathcal{O}(M)$, where $M$ denotes the total number of samples. For the contrastive loss, calculating the pairwise inner products between each hidden state and all bin centers entails a computational complexity of $\mathcal{O}(KM)$, with $K$ representing the number of bins.

\subsection{Evaluation Pipeline of NoRo}

To better evaluate NoRo, extra random noise is added to the speech feature $X$ to create noisy feature $X=Z+N+N'$ to simulate more noisy conditions, which is detailed in Section \ref{sec:noise setting}.

The whole evaluation pipeline of NoRo is formalized in Alg. \ref{alg:tesing}. Following controlled noise injection, the augmented feature $X'$ is generated through the pre-trained MLP encoder. Then, $X'$ is employed to predict UPDRS scores, where lower prediction error $E$ demonstrates higher noise robustness of UPDRS prediction.

\begin{algorithm}
	\caption{Evaluation Pipeline of NoRo}
	\label{alg:tesing}
	\small
	\begin{algorithmic}[1]
		\Require Speech Feature Matrix $X$; Pre-Trained MLP Encoder $W$; Extra Noise $N'$; Downstream Prediction Model $P$; True UPDRS Score $\bm{y}$; The Error Function $||\cdot||$
		\Ensure Prediction Errors $E$
		
		\State (Optional) Create noisy speech feature $X\leftarrow X+N'$
		\State Calculate the augmented feature $X'\leftarrow[X,\sigma(WX)]$
		\State Calculate the predicted UPDRS score $\hat{\bm{y}}\leftarrow P(X')$
        
    \Return $E\leftarrow ||\bm{y},\hat{\bm{y}}||$
	\end{algorithmic}
\end{algorithm}


\section{Experimental Settings}

\subsection{Dataset} 
\label{section:Dataset}

The real-world PD telemonitoring dataset from the Machine
Learning Repository at the University of California, Irvine (UCI) is used in this work \cite{misc_parkinsons_telemonitoring_189}, also used by other PD telemonitoring works \cite{nilashi2019analytical, nilashi2023parkinson, zhao2015householder, nilashi2023early}.

This dataset contains a total of 5875 speech test samples collected from 42 PD patients through multiple measurements. Each sample includes 2 labels, Motor UPDRS and Total UPDRS, along with 16 speech features, \{Jitter(\%)\}, \{Jitter(Abs)\}, \{Jitter:RAP\}, \{Jitter:PPQ5\}, \{Jitter:DDP\}, \{Shimmer\}, \{Shimmer(dB)\}, \{Shimmer:APQ3\}, \{Shimmer:APQ5\}, \{Shimmer:APQ11\}, \{Shimmer:DDA\}, \{NHR\}, \{HNR\}, \{RPDE\}, \{DFA\}, \{PPE\}.

The data is divided into a training set and a testing set. A 10-fold cross-validation approach is employed in this work. The split of dataset with an additional validation set is presented in Tab. \ref{tab:Dataset_Split} for different UPDRS, same as the split of \cite{zhao2015householder}. The model exhibiting the lowest loss during the validation step is preserved.

\begin{table}[htbp]
    \centering
    \caption{Dataset Split}
    \renewcommand\tabcolsep{3.8pt}
    \renewcommand{\arraystretch}{1.5}
    \scalebox{0.8}
    {
        \begin{tabular}{cccc}
            \toprule
            Label & Training & Valid & Test \\
            \midrule
            Motor UPDRS   & 2700 &300 & 2875 \\
            Total UPDRS   &  2700 &300 & 2875 \\
            \bottomrule
        \end{tabular}
        \label{tab:Dataset_Split}
    }
\end{table}





\subsection{Noise Setting}
\label{sec:noise setting}

To better evaluate NoRo, extra random Gaussian noise $N'$ is added to the speech feature $X$ to create the noisy feature $X=Z+N+N'$ to simulate more noisy environments \cite{mivule2013utilizing}, although the inherent noise $N$ remains unknown.


To generate random Gaussian noise $N'$, a mean value of ${\mu}=0$ is selected and the variance is determined based on the given signal-to-noise ratio (SNR).

SNR is expressed in decibels (dB). The higher the SNR, the less the signal is affected by noise, indicating better signal quality. The relationship between the signal and noise power is given by 

\begin{equation}
    {SNR} = 10 ~ log_{10} ~ ({\frac{\bm{P}_X}{\bm{P}_{N'}}})
\end{equation}
Here, $\bm{P}_X$ is the power of the original voice feature, $\bm{P}_{N'}$ is the power of the noise.

$\bm{P}_X$ is estimated by the following equation

\begin{equation}
    {P_{x_j}} = \frac{1}{M} \sum_{i=1}^{M} {x_{ij}^2}
\end{equation}
Here, ${P_{x_j}}$ is the power of the $j$-th feature. This equation indicates that the power of the $j$-th feature is represented as the mean of the square of the $j$-th original feature across all samples.

Because the power $\bm{P}_{N'}$ of Gaussian noise with ${\mu}=0$ equals to its variance ${\sigma}^2$, the variance of the Gaussian noise of the $j$-th feature dimension
 $\sigma_j^2$ is given by 

\begin{equation}
    \sigma_j^2 = P_{x_j} \cdot 10^{-\frac{SNR}{10}}
\end{equation}

Thus, each point of the $j$-th dimension of extra noise $N'$ is randomly sampled via 

\begin{equation}
    N'_{ij}\sim N(0, \sigma_j^2)
\end{equation}

\subsection{Evaluation}

\subsubsection{Baseline} 
\label{sec:baseline}

To evaluate the generalizability of NoRo, various regression models are employed to predict UPDRS. These regression models are referred to as \textit{downstream models}.

Among the regression models previously used for UPDRS prediction, non-ensemble models like Support Vector Regression (SVR) \cite{nilashi2019analytical,hemmerling2023monitoring,zhao2015householder}, GPR \cite{nilashi2023parkinson, hemmerling2023monitoring} and neural network (NN) models \cite{nilashi2019analytical}, ensemble learning models such as Bagging \cite{breiman1996bagging}, LightGBM \cite{ke2017lightgbm} and ANFIS ensemble method \cite{nilashi2022predicting,nilashi2019analytical,nilashi2023early} are used as downstream models. Further implementation details are reported in Appendix \ref{app:downstream}.


The baseline is the prediction error of downstream models directly using original noisy features $X$, while the result of NoRo is the prediction error using the augmented noisy features $X'$. If $X'$ achieves lower prediction error than baseline, it validates that NoRo improves the noise robustness of the downstream models.



\subsubsection{Metrics}

In this work, prediction errors are evaluated using root mean square error (RMSE), mean absolute error (MAE)\cite{nilashi2019analytical,zhao2015householder}, and median absolute error (MedianAE)\cite{zhang2024label}, as defined by Eq. (\ref{eq:Metrics}). The smaller these metrics, the better the prediction of UPDRS.

\begin{equation}
    \begin{aligned}
        &RMSE = \sqrt{\frac{1}{n}\sum_{i=1}^{n}{(y_i-\widehat{y}_i)}^2} \\
        &MAE = \frac{1}{n}\sum_{i=1}^n \left|y_i-\widehat{y}_i\right| \\
        &MedianAE = median\{\left|y_i-\widehat{y}_i\right|,i=1,2,\ldots,n\} \\ 
    \end{aligned}
    \label{eq:Metrics}
\end{equation}



Among the three metrics, RMSE is less robust and highly sensitive to outliers due to its quadratic term, while MAE and MedianAE are more robust, with MedianAE especially effective in mitigating outlier influence.

\subsection{Relative Error Estimation}

\label{sec:Repetition}

To minimize the impact of random variations, especially the random Gaussian noise, 10 repeated trials are conducted on each experimental setting. In every 10 repeated trials, different random seeds are selected, both $X$ and $X'$ are used on every random seeds.

To assess the effect of the feature augmentation method, the relative errors in every 10 repeated trials between all kinds of prediction errors based on $X$ and $X'$ are calculated. Since directly calculating the mean and standard deviation of the relative errors across 10 trials with distinct random seeds would introduce significant statistical bias, the mean and standard deviation values of the prediction errors are recorded. These statistics will be used in the estimation of the relative error's mean and standard deviation across each 10 repeated trials.

$\overline{E}_{x'}$, $s_{E_{x'}}^2$ and $\overline{E}_x$, $s_{E_x}^2$ represent the prediction error's mean and variance using $X'$ and $X$. The estimation of the mean and standard deviation of the relative error $\delta$ are calculated through Eq. (\ref{eq:mean_deviation}) and Eq. (\ref{eq:sigma_deviation}) \cite{ku1966notes}.




\begin{equation}
	\hat{\delta} = \frac{\overline{E}_{x'} - \overline{E}_x}{\overline{E}_x}
	\label{eq:mean_deviation}
\end{equation}


\begin{equation}
	\hat{\sigma}_{\delta} = \sqrt{{\left(\frac{\overline{E}_{x'}}{\overline{E}_x}\right)}^{2}\cdot
		\left(\frac{s_{E_x}^2}{{\overline{E}_x}^2}+
		\frac{{s_{E_{x'}}^2}}{{\overline{E}_{x'}}^2}\right)
	}
	\label{eq:sigma_deviation}
\end{equation}

Relative error $\hat{\delta}<0$ indicates the feature augmentation method improves the robustness of the downstream models because $\overline{E}_{x'} < \overline{E}_x$. The lower the value of $\hat{\delta}$, the greater the robustness provided by feature augmentation method. Additionally, lower $\hat{\sigma}_{\delta}$ indicates less sensitivity to randomness.

\section{Results}

The purpose of the following experiments is to address the following research questions.

RQ1: Can NoRo enhance the robustness of downstream methods against noise?

RQ2: How does NoRo perform under different SNR levels of extra noise?

RQ3: Is NoRo consistently effective across different hyperparameter settings?

RQ4: Is the feature selection module effective?

RQ5: Why the feature augmentation method is noise-robust?

\subsection{Quantitative Analysis (RQ1)} \label{sec:Quantatitive Analysis}

To evaluate the effectiveness of NoRo, downstream models are tested across different noise environments, including a non-extra noise environment and environments with extra noise at different SNR levels. The baseline and NoRo prediction errors of both Motor UPDRS and Total UPDRS are reported.

\subsubsection{Non-Extra Noise Environment}

To evaluate NoRo on the original data without extra noise $N'$ introduced. The prediction errors are shown in Tab. \ref{tab:Ground Truth}.

\begin{table*}[htbp]
  \centering
  \caption{Evaluation without extra noise. Better prediction performances (lower prediction errors) of the same downstream model between baseline and NoRo are highlighted in \textbf{bold}. Note that, because no random Gaussian noise is introduced and the hyperparameter settings of downstream models are fixed, only 1 trial is conducted in the identical experimental setting.}
    \begin{tabular}{cccccccc}
    \toprule
    \multicolumn{2}{c}{UPDRS} & \multicolumn{3}{c}{Motor UPDRS} & \multicolumn{3}{c}{Total UPDRS} \\
    \midrule
    \multicolumn{2}{c}{Error} & RMSE  & MAE   & MedianAE & RMSE  & MAE   & MedianAE \\
    \midrule
    \multirow{2}[4]{*}{SVR} & Baseline     & 1.672 & 0.836 & 0.680 & 2.267 & 0.825 & 0.618 \\
\cmidrule{2-8}          &NoRo & \textbf{1.215} & \textbf{0.794} & \textbf{0.664} & \textbf{1.575} & \textbf{0.784} & \textbf{0.595} \\
    \midrule
    \multirow{2}[4]{*}{NN} & Baseline     & 0.942 & 0.794 & 0.709 & 0.954 & 0.775 & 0.692 \\
\cmidrule{2-8}          & NoRo & \textbf{0.910} & \textbf{0.769} & \textbf{0.707} & \textbf{0.929} & \textbf{0.762} & \textbf{0.682} \\
    \midrule
    \multirow{2}[4]{*}{GPR} & Baseline     & 1.503 & 1.103 & 0.836 & 1.467 & 1.075 & 0.803 \\
\cmidrule{2-8}          & NoRo & \textbf{1.360} & \textbf{1.017} & \textbf{0.789} & \textbf{1.332} & \textbf{0.993} & \textbf{0.759} \\
    \midrule
    \multirow{2}[4]{*}{Bagging} & Baseline     & 0.845 & 0.673 & 0.574 & 0.856 & 0.660 & \textbf{0.534} \\
\cmidrule{2-8}          & NoRo & 0.845 & \textbf{0.668} & \textbf{0.555} & \textbf{0.852} & \textbf{0.658} & 0.536 \\
    \midrule
    \multirow{2}[4]{*}{LightGBM} & Baseline     & 0.824 & 0.666 & 0.568 & 0.819 & 0.644 & 0.538 \\
\cmidrule{2-8}          & NoRo & \textbf{0.823} & \textbf{0.661} & \textbf{0.567} & \textbf{0.817} & \textbf{0.638} & \textbf{0.522} \\
    \midrule
    \multirow{2}[4]{*}{ANFIS Ensemble} & Baseline     & 0.991 & 0.853 & 0.788 & 1.005 & 0.818 & \textbf{0.716} \\
\cmidrule{2-8}          & NoRo & 0.991 & \textbf{0.851} & \textbf{0.785} & 1.005 & \textbf{0.817} & 0.717 \\
    \bottomrule
    \end{tabular}%
  \label{tab:Ground Truth}%
\end{table*}%

The lower errors achieved by NoRo compared to baseline ‌demonstrate enhances the noise ‌robustness across all downstream models under this non-extra noise environment. 

However, Total UPDRS prediction shows an unexpected pattern where baseline yields lower MedianAE compared to NoRo on both Bagging and ANFIS Ensemble methods. NoRo is specifically designed for noisy environments, while the current non-extra noise environment contains minimal noise. This low-noise environment creates suboptimal operating parameters for NoRo, which may result in higher prediction errors.


\subsubsection{More Noisy Environments}

To comprehensively evaluate the effectiveness of NoRo, downstream models are tested under more noisy environments with ‌extra noise at SNR=10, 20, 30dB‌. Baseline and NoRo prediction errors are shown in Tab. \ref{tab:Quantatitive Analysis}.


\begin{table*}
    \centering
    \caption{Evaluation with extra noise. Better prediction performances under the same condition between baseline and NoRo are highlighted in \textbf{bold}. Statistically significant differences $(p < 0.05)$ observed in the 10 repeated trials are marked with an asterisk (*).}
\begin{tabular}{cccccccc}
    \toprule
    \multicolumn{1}{c}{Motor UPDRS} & Error & Baseline     & NoRo & Baseline    & NoRo & Baseline     & NoRo \\
    \midrule
    Noise & Models & \multicolumn{2}{c}{SVR} & \multicolumn{2}{c}{NN} & \multicolumn{2}{c}{GPR} \\
    \midrule
    \multirow{3}[6]{*}{10dB} & RMSE  & $9.587_{\pm0.503}$ & $\mathbf{4.728}_{\pm 0.217}^*$ & $1.259_{\pm0.009}$ & $\mathbf{1.136}_{\pm0.011}^*$ & $1.984_{\pm0.080}$ & $\mathbf{1.435}_{\pm 0.037}^*$ \\
\cmidrule{2-8}         & MAE   & $4.641_{\pm0.161}$ & $\mathbf{2.802}_{\pm0.062}^*$ & $1.006_{\pm0.006}$  & $\mathbf{0.934}_{\pm0.010}^*$ & $1.313_{\pm0.022}$  & $\mathbf{1.061}_{\pm0.013}^*$ \\
\cmidrule{2-8}          & MedianAE & $1.799_{\pm0.071}$ & $\mathbf{1.655}_{\pm0.054}^*$ & $0.858_{\pm0.011}$ & $\mathbf{0.830}_{\pm0.012}^*$ & $0.942_{\pm0.015}$ & $\mathbf{0.861}_{\pm0.007}^*$ \\
    \midrule
    \multirow{3}[6]{*}{20dB} & RMSE & $1.944_{\pm0.124}$ & $\mathbf{1.494}_{\pm0.059}^*$ & $0.983_{\pm0.003}$ & $\mathbf{0.961}_{\pm0.005}^*$ & $3.185_{\pm0.042}$ & $\mathbf{2.458}_{\pm0.038}^*$ \\
\cmidrule{2-8}          & MAE & $1.003_{\pm0.016}$ & $\mathbf{0.960}_{\pm0.010}^*$ & $0.819_{\pm0.002}$ & $\mathbf{0.806}_{\pm0.003}^*$ & $2.238_{\pm0.025}$ & $\mathbf{1.750}_{\pm0.026}^*$ \\
\cmidrule{2-8}  & MedianAE & $0.790_{\pm0.013}$ & $\mathbf{0.779}_{\pm0.012}^*$ & $\mathbf{0.732}_{\pm0.008}$ & $0.736_{\pm0.006}$ & $1.524_{\pm0.022}$ & $\mathbf{1.228}_{\pm0.029}^*$ \\
    \midrule
    \multirow{3}[6]{*}{30dB} & RMSE  & $1.703_{\pm0.046}$ &$\mathbf{1.244}_{\pm0.019}^*$ & $0.947_{\pm0.001}$ & $\mathbf{0.918}_{\pm0.002}^*$ & $1.818_{\pm0.039}$ &$\mathbf{1.587}_{\pm0.023}^*$ \\
\cmidrule{2-8}          & MAE   & $0.848_{\pm0.004}$ & $\mathbf{0.807}_{\pm0.004}^*$ & $0.797_{\pm0.001}$ & $\mathbf{0.774}_{\pm0.002}^*$ & $1.328_{\pm0.023}$ & $\mathbf{1.185}_{\pm0.014}^*$ \\
\cmidrule{2-8}          & MedianAE & $0.689_{\pm0.006}$ & $\mathbf{0.676}_{\pm0.006}^*$ & $0.717_{\pm0.003}$ & $\mathbf{0.711}_{\pm0.005}^*$ & $1.000_{\pm0.020}$ & $\mathbf{0.913}_{\pm0.025}^*$ \\
\midrule  
Noise & Models & \multicolumn{2}{c}{Bagging} & \multicolumn{2}{c}{LightGBM} & \multicolumn{2}{c}{ANFIS Ensemble} \\
\midrule
\multirow{3}[6]{*}{10dB} & RMSE  & $1.074_{\pm0.009}$ &$ \mathbf{1.073}_{\pm0.010} $&$ \mathbf{1.145}_{\pm0.015} $&$ 1.146_{\pm0.014} $&$ 0.994_{\pm0.001} $&$ \mathbf{0.994}_{\pm0.000}^*$ \\
\cmidrule{2-8}          & MAE   &$ 0.883_{\pm0.009} $&$ \mathbf{0.883}_{\pm0.008} $&$ 0.924_{\pm0.011} $&$ \mathbf{0.923}_{\pm0.011} $&$ 0.854_{\pm0.001} $&$ \mathbf{0.854}_{\pm0.000}^*$ \\
\cmidrule{2-8}          & MedianAE & $0.794_{\pm0.018}$&$ \mathbf{0.793}_{\pm0.012} $&$ 0.796_{\pm0.014} $&$ \mathbf{0.795}_{\pm0.017} $&$ 0.782_{\pm0.000} $&$ \mathbf{0.779}_{\pm0.000}^*$ \\
\midrule
\multirow{3}[6]{*}{20dB} & RMSE  &$ 0.985_{\pm0.010} $&$ \mathbf{0.983}_{\pm0.011} $&$ 0.976_{\pm0.011} $&$ \mathbf{0.974}_{\pm0.009} $&$ {0.993}_{\pm0.002} $&$ \mathbf{0.992}_{\pm0.001}$ \\
\cmidrule{2-8}          & MAE   &$ 0.800_{\pm0.008} $&$ \mathbf{0.799}_{\pm0.009} $&$ 0.786_{\pm0.008} $&$ \mathbf{0.785}_{\pm0.007} $&$ 0.853_{\pm0.002} $&$ \mathbf{0.852}_{\pm0.000}^*$ \\
\cmidrule{2-8}          & MedianAE &$ 0.707_{\pm0.010} $&$ \mathbf{0.700}_{\pm0.012}^*$&$ \mathbf{0.676}_{\pm0.010} $&$ 0.678_{\pm0.009} $&$ {0.787}_{\pm0.005} $&$ \mathbf{0.782}_{\pm0.001}^*$ \\
\midrule
\multirow{3}[6]{*}{30dB} & RMSE  &$ 0.877_{\pm0.006} $&$ \mathbf{0.876}_{\pm0.006} $&$ \mathbf{0.851}_{\pm0.006} $&$ 0.852_{\pm0.006} $&$ \mathbf{0.991}_{\pm0.001} $&$ 0.991_{\pm0.000}$ \\
\cmidrule{2-8}          & MAE   &$ 0.702_{\pm0.004} $&$ \mathbf{0.700}_{\pm0.005} $&$ \mathbf{0.685}_{\pm0.004} $&$ 0.686_{\pm0.004} $&$ 0.853_{\pm0.001} $&$ \mathbf{0.851}_{\pm0.000} $\\
\cmidrule{2-8}          & MedianAE &$ 0.605_{\pm0.009} $&$ \mathbf{0.598}_{\pm0.011} $&$ 0.590_{\pm0.006} $&$ \mathbf{0.588}_{\pm0.006} $&$ {0.787}_{\pm0.003} $&$ \mathbf{0.784}_{\pm0.001}^*$ \\
\hline\hline  
    Total UPDRS & Models & \multicolumn{2}{c}{SVR} & \multicolumn{2}{c}{NN} & \multicolumn{2}{c}{GPR} \\
    \midrule
    \multirow{3}[6]{*}{10dB} & RMSE  &$ 9.173_{\pm0.486} $&$ \mathbf{3.963}_{\pm0.157}^* $&$ 1.233_{\pm0.008} $&$ \mathbf{1.128}_{\pm0.010}^* $&$ 1.946_{\pm0.071} $&$ \mathbf{1.435}_{\pm0.041}^*$ \\
\cmidrule{2-8}          & MAE   &$ 4.279_{\pm0.164 }$&$ \mathbf{2.441}_{\pm0.063}^* $&$ 0.977_{\pm0.007} $&$ \mathbf{0.923}_{\pm0.009}^* $&$ 1.275_{\pm0.022} $&$ \mathbf{1.029}_{\pm0.016}^*$ \\
\cmidrule{2-8}          & MedianAE &$ 1.670_{\pm0.069} $&$ \mathbf{1.527}_{\pm0.049}^* $&$ 0.820_{\pm0.011} $&$ \mathbf{0.816}_{\pm0.019} $&$ 0.911_{\pm0.014} $&$ \mathbf{0.828}_{\pm0.013}^*$ \\
    \midrule
    \multirow{3}[6]{*}{20dB} & RMSE  &$ 2.581_{\pm0.130} $&$ \mathbf{1.724}_{\pm0.090}^* $&$ 0.988_{\pm0.003} $&$ \mathbf{0.972}_{\pm0.004}^* $&$ 3.110_{\pm0.036} $&$ \mathbf{2.419}_{\pm0.030}^*$ \\
\cmidrule{2-8}          & MAE   &$ 0.977_{\pm0.014} $&$ \mathbf{0.926}_{\pm0.011}^* $&$ 0.798_{\pm0.002} $&$ \mathbf{0.795}_{\pm0.003}^* $&$ 2.187_{\pm0.023} $&$ \mathbf{1.719}_{\pm0.018}^*$ \\
\cmidrule{2-8}          & MedianAE &$ 0.727_{\pm0.015} $&$ \mathbf{0.722}_{\pm0.014}^* $&$ 0.701_{\pm0.008} $&$ \mathbf{0.700}_{\pm0.005} $&$ 1.490_{\pm0.027} $&$ \mathbf{1.210}_{\pm0.024}^*$ \\
    \midrule
    \multirow{3}[6]{*}{30dB} & RMSE  &$ 2.329_{\pm0.043} $&$ \mathbf{1.589}_{\pm0.031}^* $&$ 0.958_{\pm0.001} $&$ \mathbf{0.935}_{\pm0.002}^* $&$ 1.769_{\pm0.034} $&$ \mathbf{1.561}_{\pm0.018}^*$ \\
\cmidrule{2-8}          & MAE   &$ 0.837_{\pm0.003} $&$ \mathbf{0.796}_{\pm0.004}^* $&$ 0.777_{\pm0.001} $&$ \mathbf{0.766}_{\pm0.001}^* $&$ 1.290_{\pm0.022} $&$ \mathbf{1.160}_{\pm0.015}^*$ \\
\cmidrule{2-8}          & MedianAE &$ 0.627_{\pm0.007} $&$ \mathbf{0.608}_{\pm0.008}^* $&$ 0.690_{\pm0.004} $&$ \mathbf{0.680}_{\pm0.006}^* $&$ 0.972_{\pm0.028} $&$ \mathbf{0.887}_{\pm0.023}^*$ \\
    \midrule
    Noise & Models & \multicolumn{2}{c}{Bagging} & \multicolumn{2}{c}{LightGBM} & \multicolumn{2}{c}{ANFIS Ensemble} \\
    \midrule
    \multirow{3}[6]{*}{10dB} & RMSE  &$ 1.137_{\pm0.013} $&$ \mathbf{1.136}_{\pm0.014} $&$ \mathbf{1.211}_{\pm0.014} $&$ 1.213_{\pm0.014} $&$ \mathbf{1.008}_{\pm0.001} $&$ 1.009_{\pm0.000}$ \\
\cmidrule{2-8}          & MAE   &$ 0.910_{\pm0.009} $&$ \mathbf{0.908}_{\pm0.010} $&$ 0.960_{\pm0.011} $&$ \mathbf{0.959}_{\pm0.012} $&$ 0.820_{\pm0.001} $&$ \mathbf{0.820}_{\pm0.000}$ \\
\cmidrule{2-8}          & MedianAE &$ 0.779_{\pm0.010} $&$ \mathbf{0.769}_{\pm0.011}^* $&$ 0.798_{\pm0.015} $&$ \mathbf{0.796}_{\pm0.013} $&$ {0.720}_{\pm0.006} $&$ \mathbf{0.720}_{\pm0.001}$ \\
    \midrule
    \multirow{3}[6]{*}{20dB} & RMSE  &$ 1.019_{\pm0.010} $&$ \mathbf{1.019}_{\pm0.009} $&$ 0.999_{\pm0.009} $&$ \mathbf{0.999}_{\pm0.009} $&$ \mathbf{1.005}_{\pm0.001} $&$ 1.006_{\pm0.000} $\\
\cmidrule{2-8}          & MAE   &$ 0.798_{\pm0.009} $&$ \mathbf{0.798}_{\pm0.006} $&$ 0.784_{\pm0.007} $&$ \mathbf{0.783}_{\pm0.008} $&$ {0.818}_{\pm0.001} $&$ \mathbf{0.818}_{\pm0.000} $\\
\cmidrule{2-8}          & MedianAE &$ \mathbf{0.652}_{\pm0.009} $&$ 0.653_{\pm0.010} $&$ 0.643_{\pm0.013} $&$ \mathbf{0.635}_{\pm0.008}^* $&$ 0.717_{\pm0.002} $&$ \mathbf{0.716}_{\pm0.003}$ \\
    \midrule
    \multirow{3}[6]{*}{30dB} & RMSE  &$ 0.901_{\pm0.007} $&$ \mathbf{0.897}_{\pm0.005}^* $&$ 0.858_{\pm0.007} $&$ \mathbf{0.856}_{\pm0.007}^* $&$ {1.005}_{\pm0.000}$&$ 1.005_{\pm0.000}$ \\
\cmidrule{2-8}          & MAE   &  $ 0.696_{\pm0.006} $&$ \mathbf{0.692}_{\pm0.004}^* $&$ 0.673_{\pm0.006} $&$ \mathbf{0.669}_{\pm0.005}^* $&$ {0.818}_{\pm0.000} $&$ \mathbf{0.817}_{\pm0.000}$ \\
\cmidrule{2-8}          & MedianAE &$ \mathbf{0.557}_{\pm0.010} $&$ 0.558_{\pm0.009} $&$ 0.549_{\pm0.009} $&$ \mathbf{0.544}_{\pm0.009} $&$ 0.717_{\pm0.003} $&$ \mathbf{0.717}_{\pm0.002}$ \\
    \end{tabular}
    \label{tab:Quantatitive Analysis}%
  \end{table*}%

First, compare all Baseline columns with NoRo columns, most prediction errors using NoRo are significantly lower than baseline, while the prediction errors with NoRo higher than baseline are not significant (especially LightGBM and ANFIS Ensemble). Thus, NoRo enhances the noise robustness of nearly all downstream models.
    
Second, compare the results of non-ensemble models (SVR, NN, GPR) with ensemble models (Bagging, LightGBM, ANFIS Ensemble), NoRo significantly enhances the robustness of non-ensemble models but has limited impact on ensemble models. Because ensemble models integrate the prediction from many submodels by averaging or voting, wild prediction errors caused by noise are reduced. Thus, ensemble models have inherent robustness against noise where NoRo exhibits subtle impact, or even unexpected but insignificant prediction error increase as mentioned earlier.


To conclude, NoRo demonstrates noise robustness on nearly all downstream models in different noise environment (without or with extra noise at different SNR levels), which is more significant on non-ensemble methods. 

\subsection{Qualitative Analysis (RQ2)}

\begin{figure*}[htbp]
    \centering
        \subfigure[Motor UPDRS RMSE]
        {\includegraphics[width=0.32\textwidth]{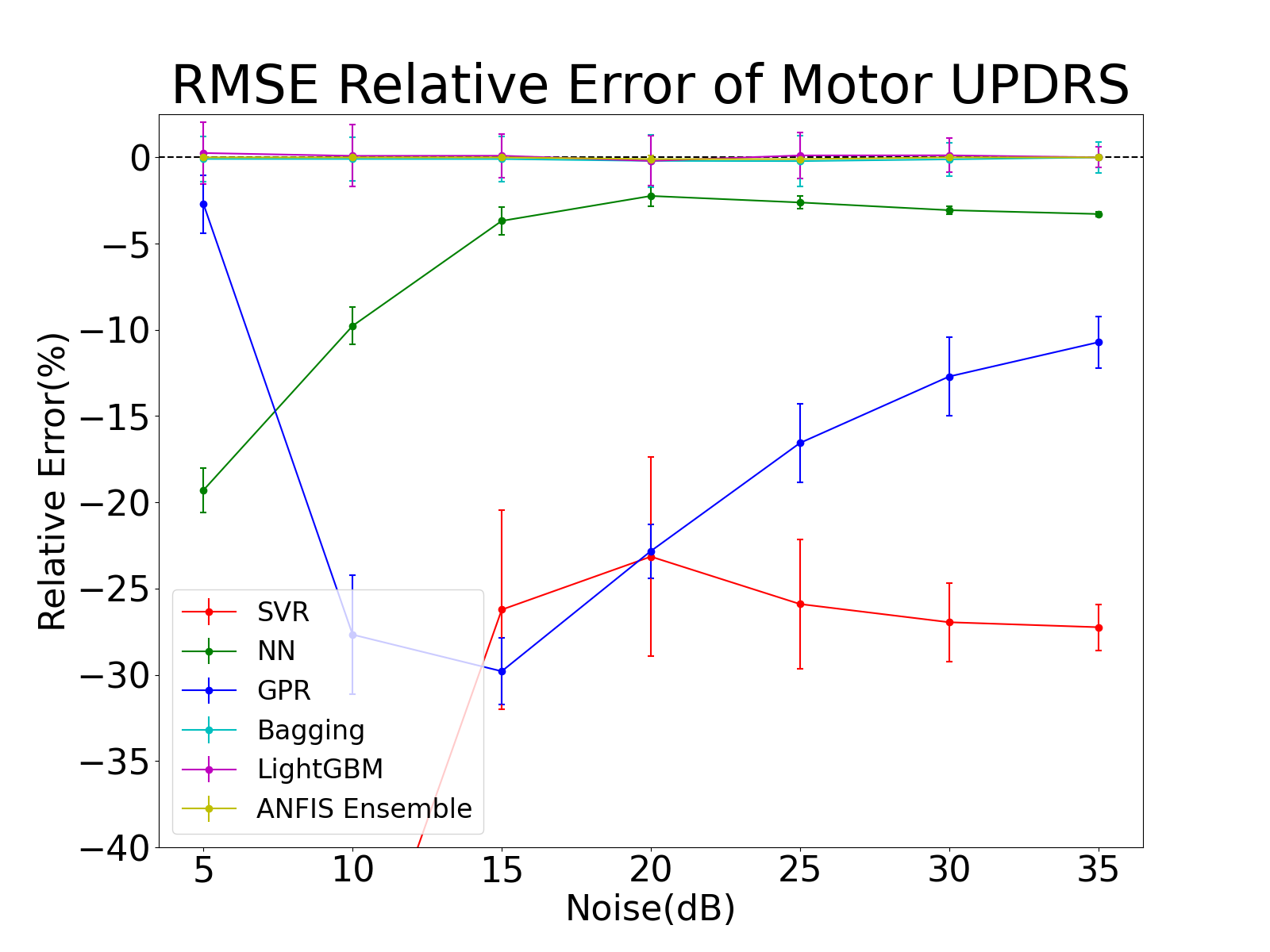}
        \label{fig:4.a}}
        \subfigure[Motor UPDRS MAE]
        {\includegraphics[width=0.32\textwidth]{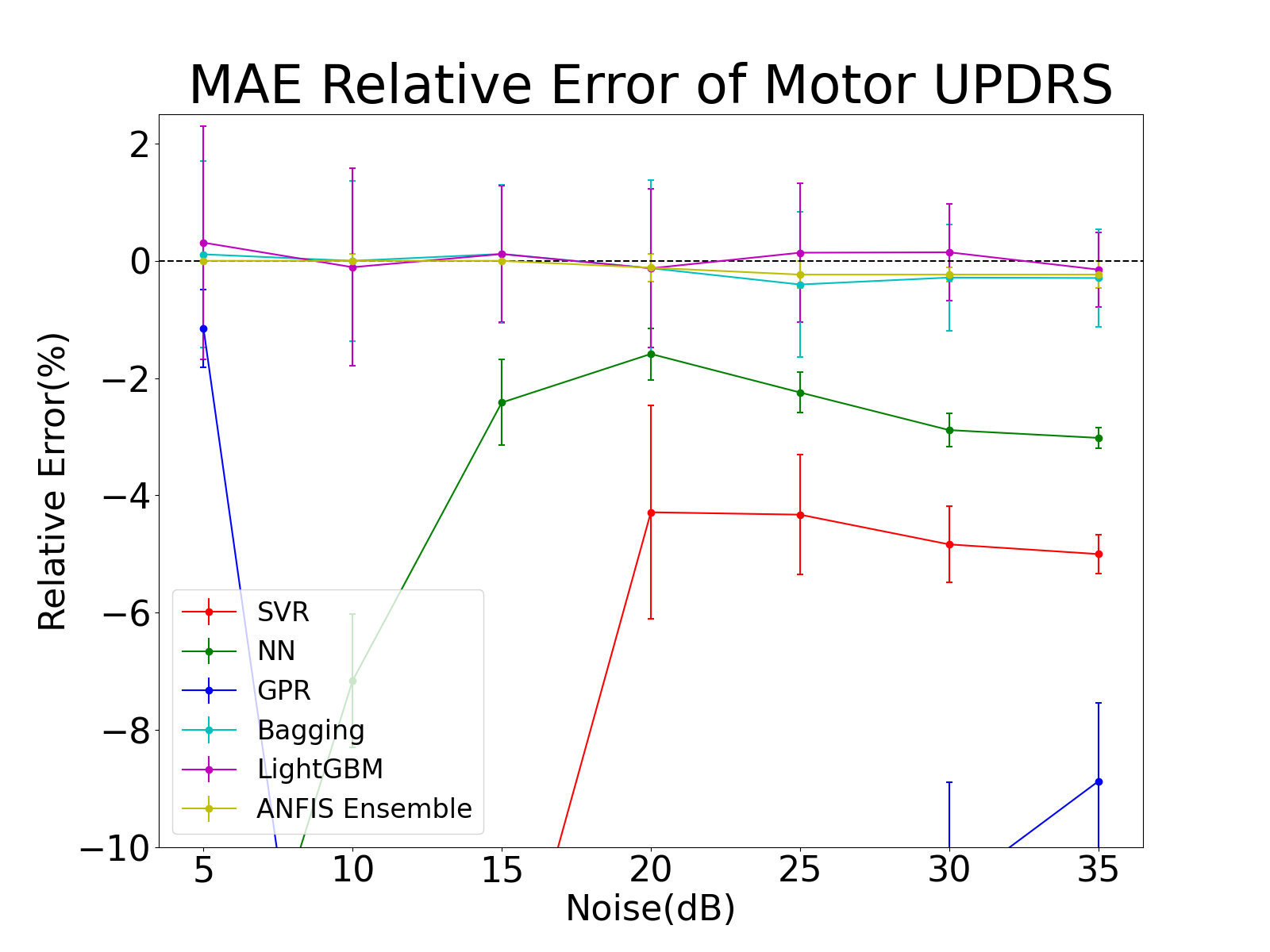}
        \label{fig:4.b}}
        \subfigure[Motor UPDRS MedianAE]
        {\includegraphics[width=0.32\textwidth]{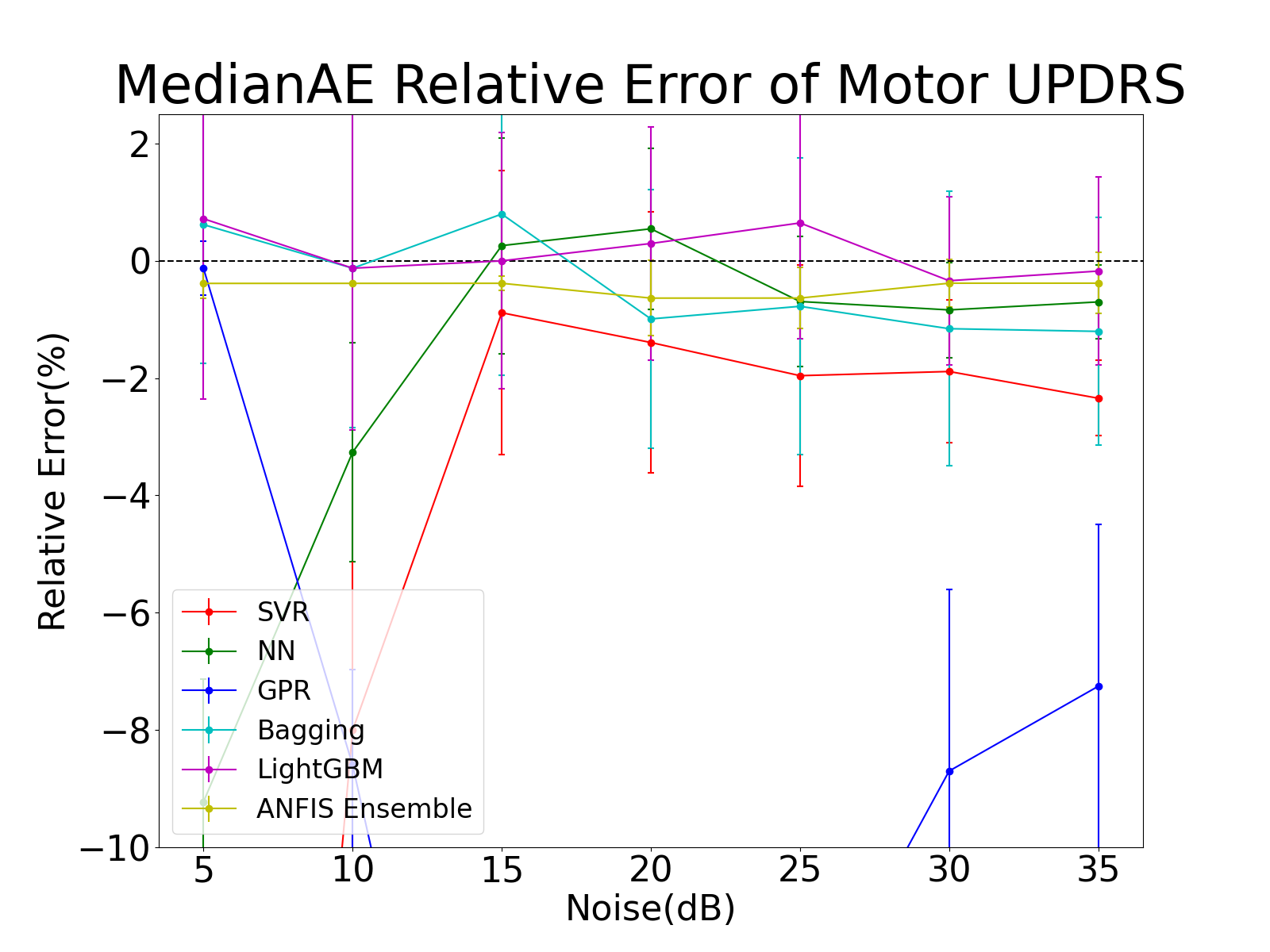}
        \label{fig:4.c}}

        \subfigure[Total UPDRS RMSE]
        {\includegraphics[width=0.32\textwidth]{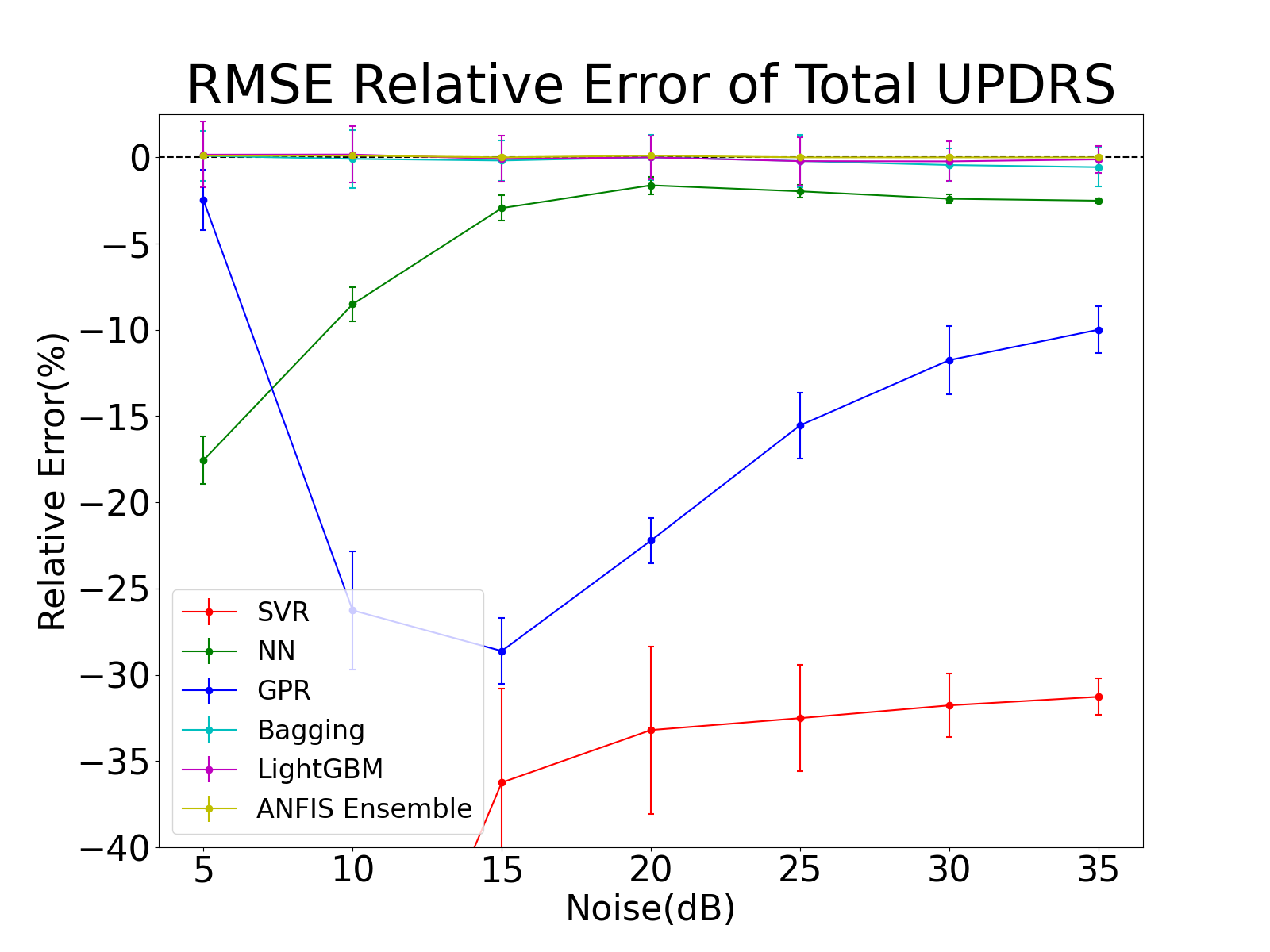}
        \label{fig:4.d}}
        \subfigure[Total UPDRS MAE]
        {\includegraphics[width=0.32\textwidth]{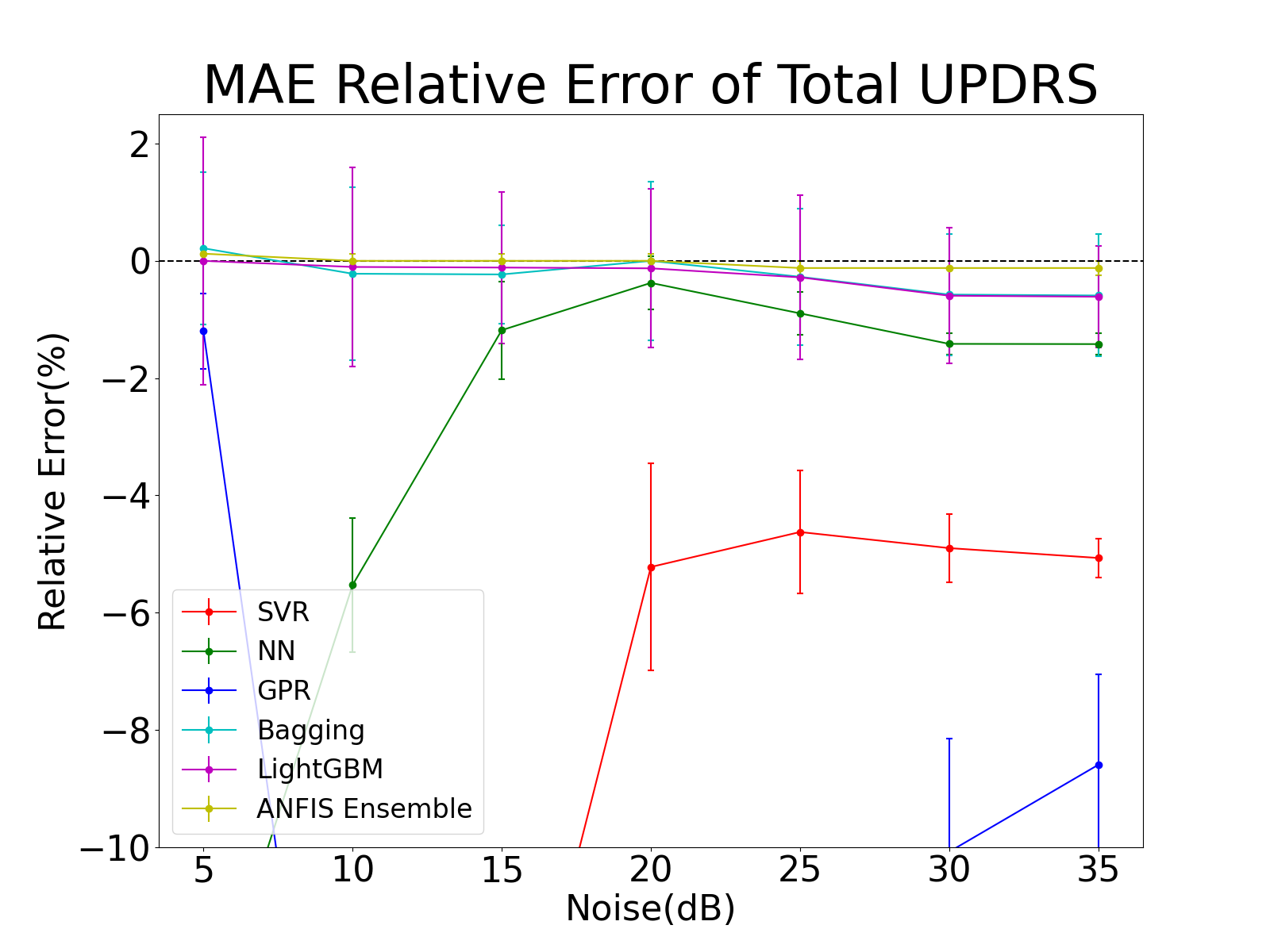}
        \label{fig:4.e}}
        \subfigure[Total UPDRS MedianAE]
        {\includegraphics[width=0.32\textwidth]{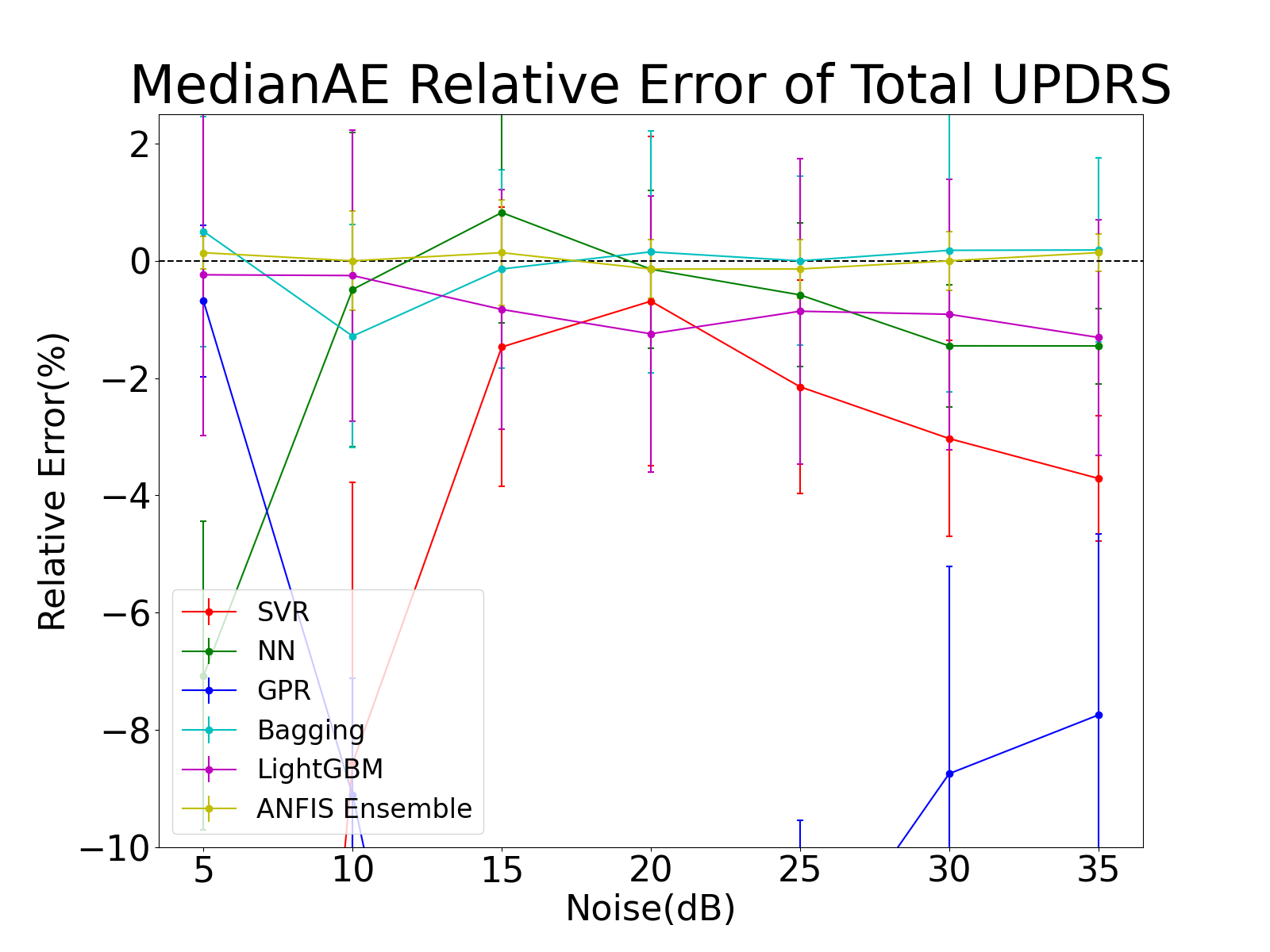}
        \label{fig:4.f}}
    \caption{Qualitative Analysis Results. The relative errors between the prediction errors (RMSE, MAE, MedianAE) of NoRo and baseline ($\delta=({E}_{x'} - {E}_x)/{{E}_x}$) under different SNR levels of extra noise environments are presented. Relative error $\delta<0$ demonstrates the prediction error using NoRo is better (lower) than baseline where NoRo enhances the robustness of the downstream model under the noisy environments with extra noise at certain SNR levels.}
    \label{fig:Qualitative Analysis}
\end{figure*}

To evaluate the performance of NoRo across different extra noise at different SNR levels, the relative errors of RMSE, MAE and MedianAE (detailed in Section \ref{sec:Repetition}) with extra noise at each SNR level are illustrated in Fig. \ref{fig:Qualitative Analysis}.

First, in Fig. \ref{fig:4.a} and Fig. \ref{fig:4.d}, NoRo reduces the RMSE of SVR by over 40\% when SNR=5/10dB. In other subplots in Fig. \ref{fig:Qualitative Analysis}, NoRo reduces the MAE and MedianAE of different downstream models by up to over 10\%. The 40\% reduction of RMSE is much higher than the reduction of MAE and MedianAE because the calculation method of RMSE is not noise-robust, where the quadratic term is easily influenced by noise.
    
Second, in Fig. \ref{fig:4.c}, relative errors of MedianAE for certain downstream models (e.g., Bagging and LighGBM) exhibit marginal values above $0$ yet remaining below 1\%. The subtle variation can be attributed to the inherent characteristics of the MedianAE calculation. While SNR level remains constant, the actual noise intensity varies across individual samples. Notably, MedianAE measures the absolute prediction error for samples with relatively lower extra noise levels. However, NoRo reduces prediction errors for samples of higher extra noise levels but has limited impact on lower-noise samples, resulting in stable improvement for RMSE and MAE but unstable performance for MedianAE. 


To conclude, NoRo enhances the noise robustness of most downstream models under extra noise at different SNR levels. It achieves a reduction up to more than 40\% on RMSE, and up to more than 10\% on MAE and MedianAE.


\subsection{Hyperparameter Analysis (RQ3)}\label{sec:hyperparam sensitivity}

\begin{figure*}[htbp]
    \centering
        \subfigure[Motor UPDRS RMSE]
        {\includegraphics[width=0.32\textwidth]{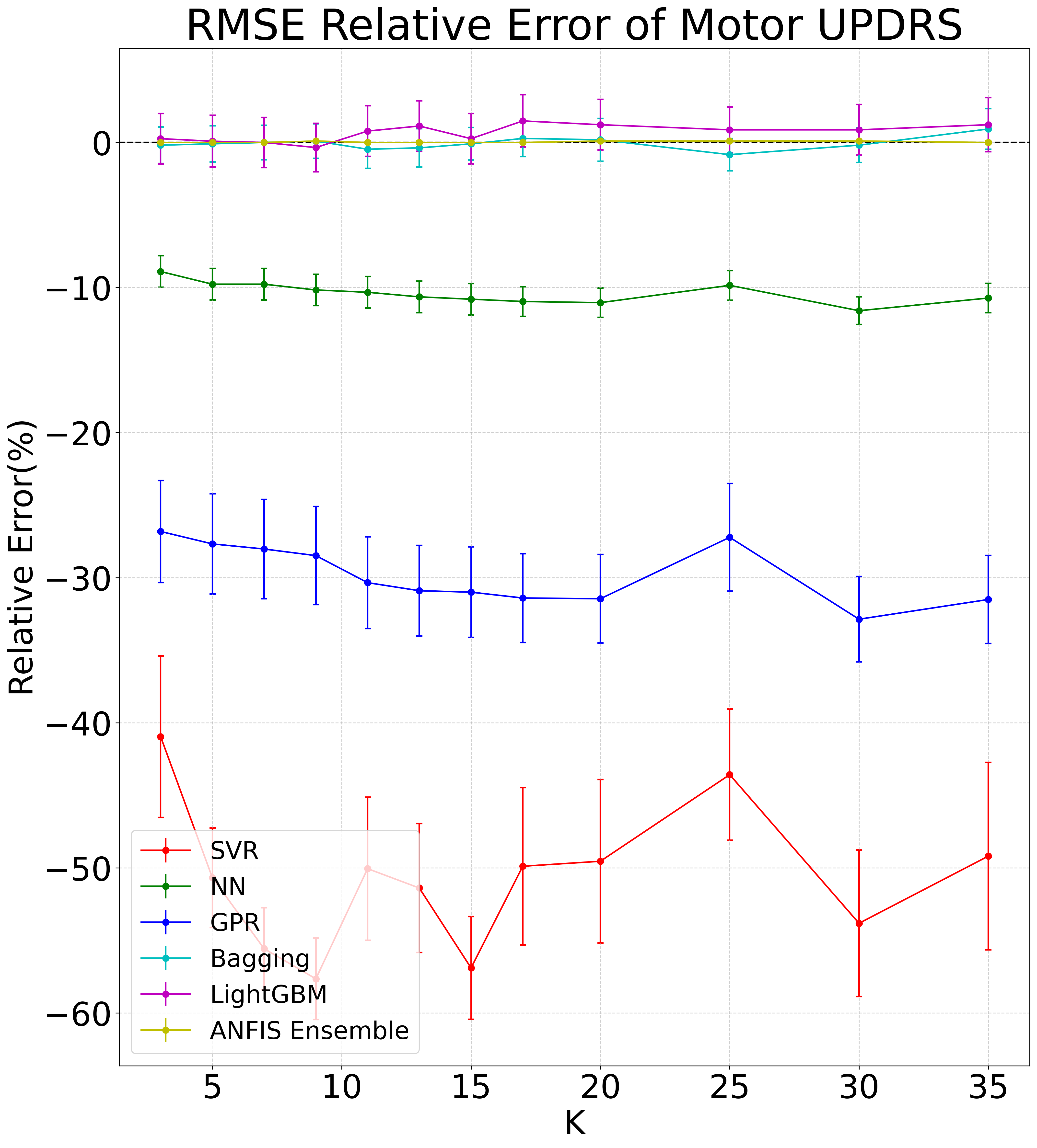}
        \label{subfig:5.a}}
        \subfigure[Motor UPDRS MAE]
        {\includegraphics[width=0.32\textwidth]{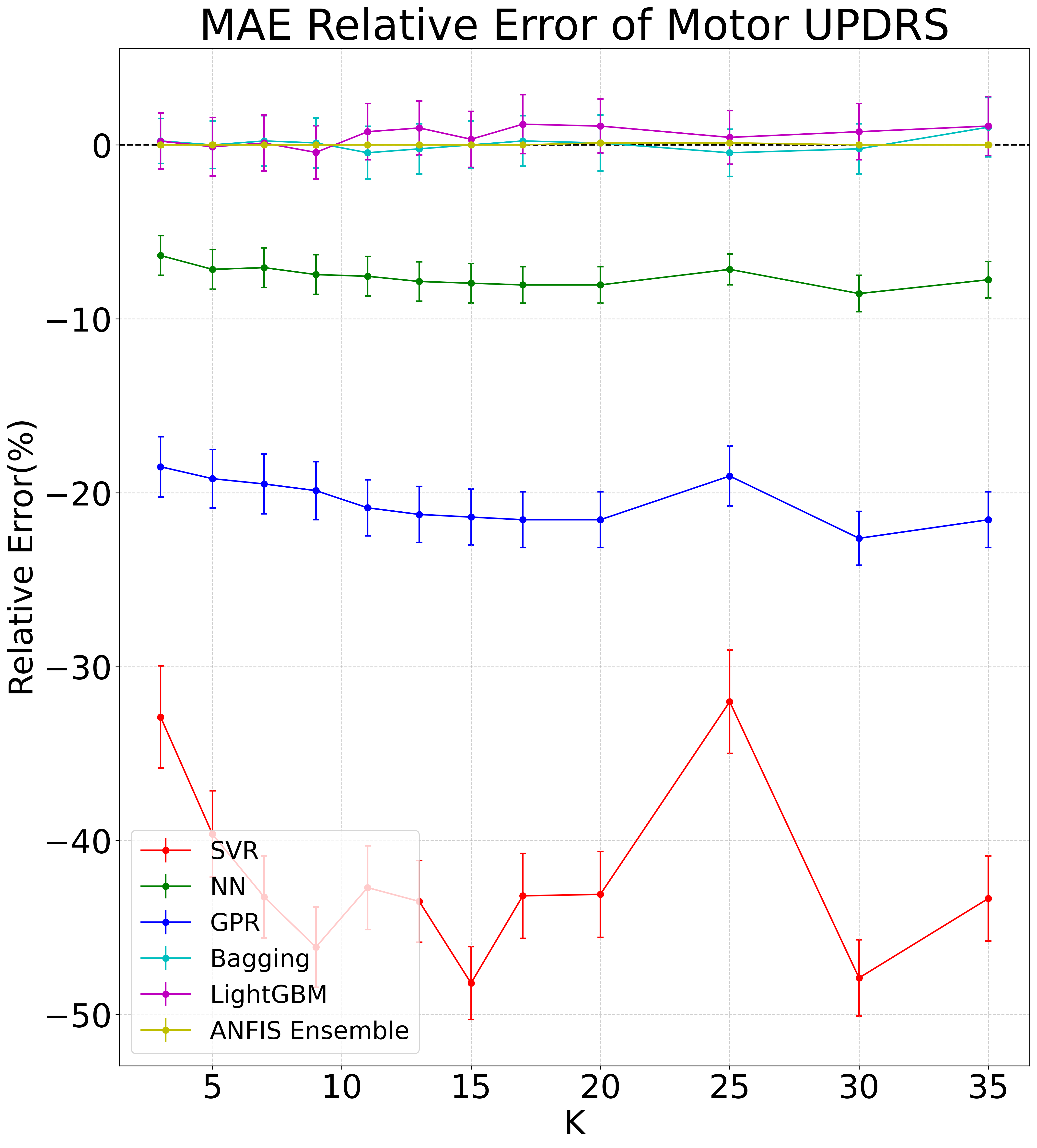}
        \label{subfig:5.b}}
        \subfigure[Motor UPDRS MedianAE]
        {\includegraphics[width=0.32\textwidth]{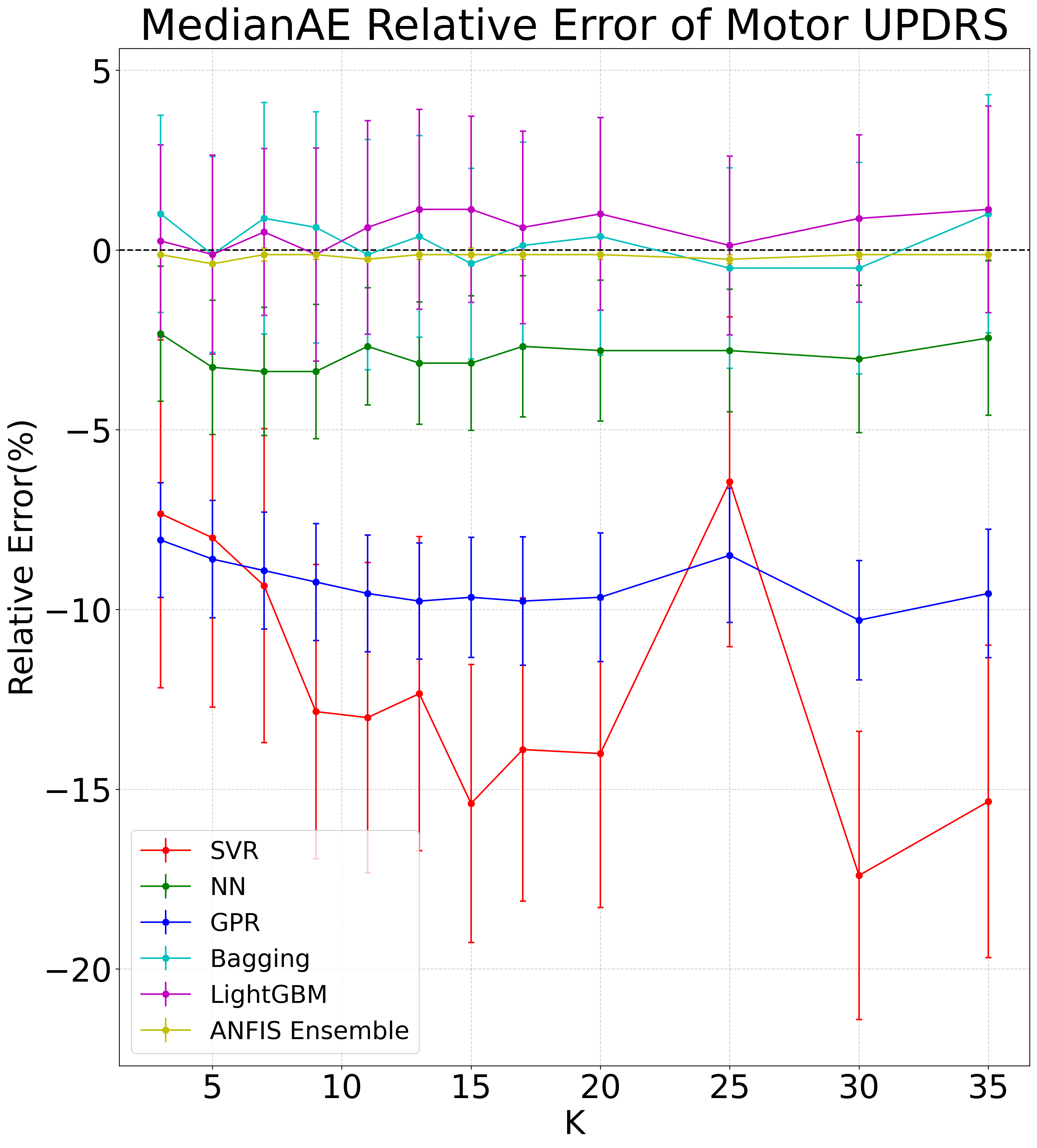}\label{subfig:5.c}}

        \subfigure[Total UPDRS RMSE]
        {\includegraphics[width=0.32\textwidth]{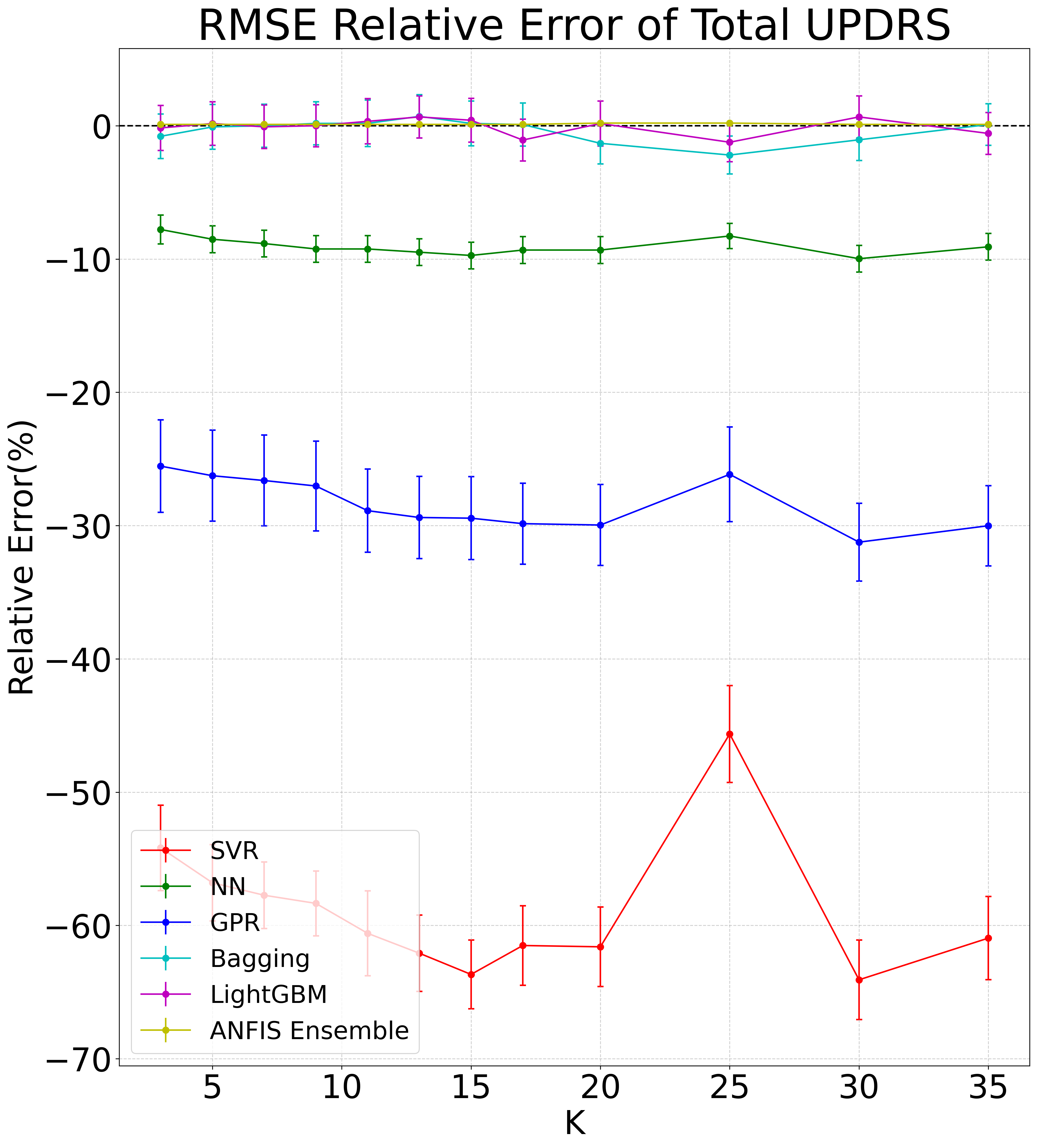}\label{subfig:5.d}}
        \subfigure[Total UPDRS MAE]
        {\includegraphics[width=0.32\textwidth]{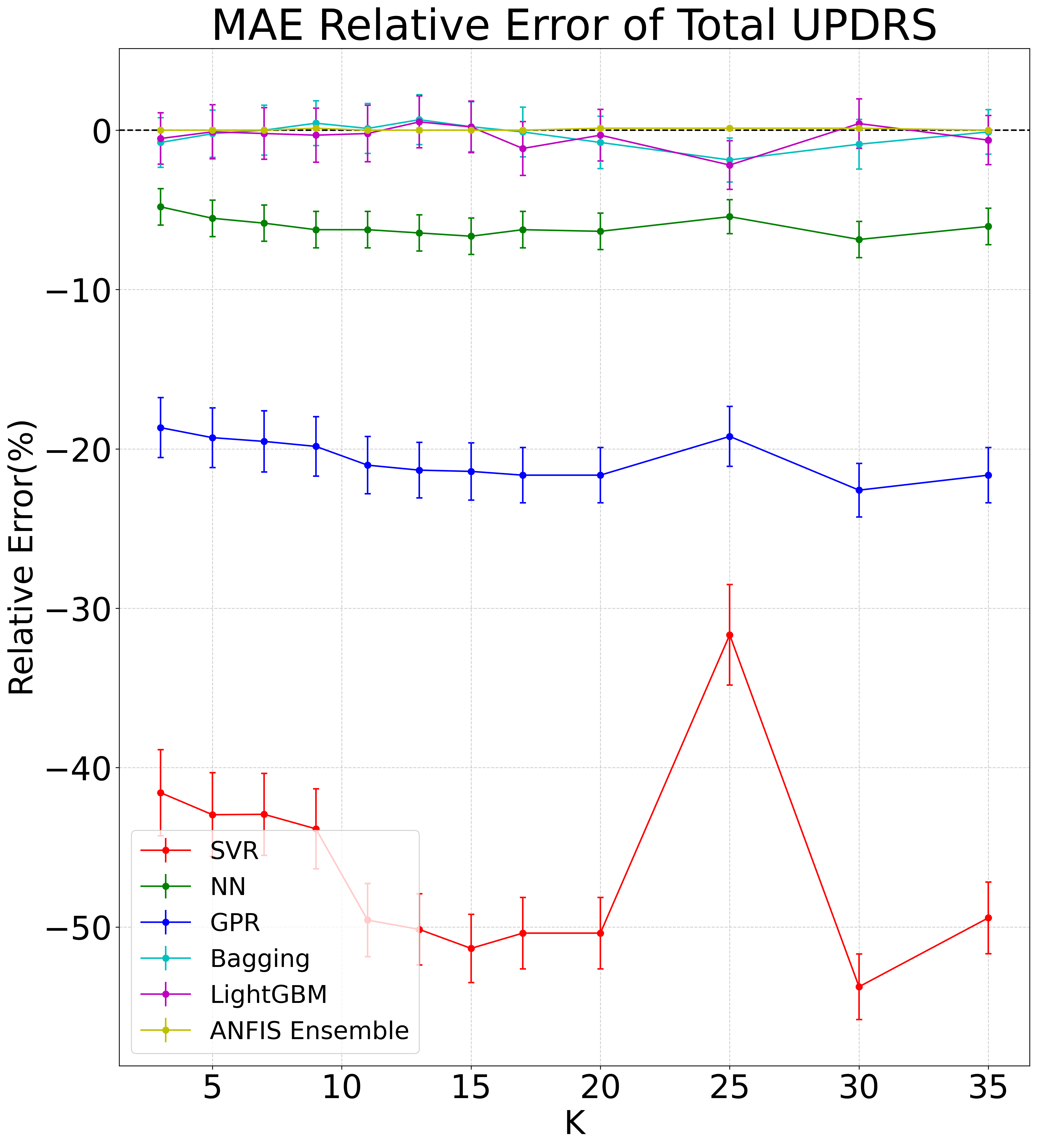}
        \label{subfig:5.e}}
        \subfigure[Total UPDRS MedianAE]
        {\includegraphics[width=0.32\textwidth]{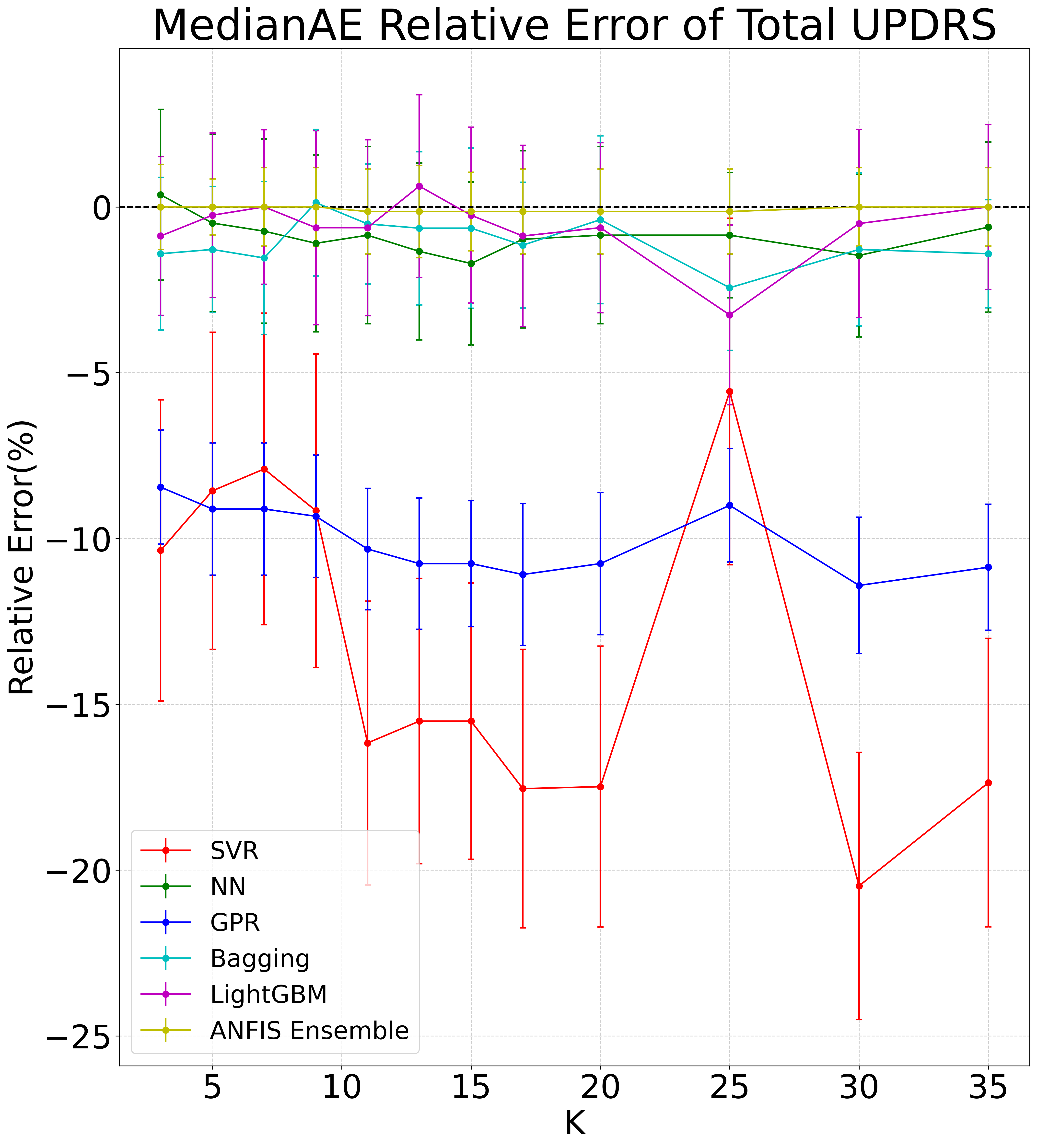}\label{subfig:5.f}}



    \caption{Results on different bin numbers $K$ settings. The relative errors of RMSE, MAE, and MedianAE between baseline and NoRo of different downstream models with extra noise at SNR=10dB are presented. Here, baseline is obtained from different MLP encoders of different bin numbers $K$ in each plot. Relative error < 0  demonstrates the prediction error using NoRo is better (lower) than baseline, then NoRo with the certain hyperparameter $K$ enhances the robustness of downstream models.}
    \label{fig:Bin Numer}
\end{figure*}

To evaluate the performance of NoRo across different hyperparameter settings, different MLP encoders are trained with different bin number $K$s, and are tested under noisy environment with extra noise at SNR=10dB. The results are shown in Fig. \ref{fig:Bin Numer}.


First, in all subplots of Fig. \ref{fig:Bin Numer}, relative errors exhibit remarkable stability across various $K$ values for all downstream models except SVR, whose regression robustness is weaker. Thus, for nearly all downstream models, NoRo exhibits hyperparameter robustness across nearly all $K$ settings. However, at $K=25$, nearly all curves demonstrate unexpected severe deviations. This phenomenon appears to come from the distinct binning pattern specific to the $K=25$ setting. 

Second, in all subplots of Fig. \ref{fig:Bin Numer}, for SVR, NN and GPR, with the increase of $K$, the relative errors decrease. As the bin number $K$ increases, samples are partitioned into more fine-grained bins, which enhances robustness against noise interference. Despite larger $K$ settings introduce higher computational demands, these $K$ settings simultaneously deliver better performance outcomes.


Third, compare Fig. \ref{subfig:5.a}-Fig. \ref{subfig:5.c} with Fig. \ref{subfig:5.d}-Fig. \ref{subfig:5.f}, the relative errors of Total UPRDS prediction are lower than Motor UPDRS in general using ensemble learning methods (Bagging, LightGBM, ANFIS Ensemble). This phenomenon stems from Total UPDRS's inherent complexity as a comprehensive metric integrating Motor UPDRS with other UPDRS scores. Notably, ensemble learning methods can simultaneously leverage different components of Total UPDRS. Thus, these downstream models are better at Total UPDRS prediction than Motor UPDRS alone.

To conclude, NoRo is barely sensitive to $K$ settings ($K=25$ is an exception) across all downstream models except the least robust regression method SVR. With the increase of $K$, the effectiveness of NoRo increases.

\subsection{Effectiveness of Feature Selection Module (RQ4)}

To validate the effectiveness of the feature selection module detailed in Section \ref{sec:feature selection}, \{Jitter:RAP\}, the feature with the lowest importance is employed as the binning feature. The prediction errors are tested under $K$=5, SNR=20dB in Tab. \ref{tab:Binning Feature}.

\begin{table*}[htbp]
    \centering
    \caption{Effectiveness of the feature selection module. For the prediction errors using NoRo, $X'$ is obtained from the MLP encoder trained with the binning feature \{Jitter:RAP\} ranking the lowest importance score. Compared with the SNR = 20dB rows of Tab. \ref{tab:Quantatitive Analysis} using the binning feature \{DFA\} ranking the highest importance score, the prediction errors of NoRo obviously increase.}
      \begin{tabular}{ccccccc}
      \toprule
            & Baseline     & NoRo & Baseline    & NoRo & Baseline     & NoRo \\
      \midrule
      Motor UPDRS & \multicolumn{2}{c}{RMSE} & \multicolumn{2}{c}{MAE} & \multicolumn{2}{c}{MedianAE} \\
      \midrule
      SVR   & $1.944_{\pm0.124}$ & $\mathbf{1.487}_{\pm0.077}^*$ & $1.003_{\pm0.016}$ & $\mathbf{0.925}_{\pm0.012}^*$ & ${0.790}_{\pm0.013}$ & $\mathbf{0.745}_{\pm0.008}^*$ \\
      \midrule
      NN    & $0.983_{\pm0.003}$ & $\mathbf{0.970}_{\pm0.004}^*$ & $0.819_{\pm0.002}$ & $\mathbf{0.813}_{\pm0.003}^*$ & $\mathbf{0.732}_{\pm0.008}^*$ & $0.737_{\pm0.010}$ \\
      \midrule
      GPR   & $3.185_{\pm0.042}$ & $\mathbf{3.156}_{\pm0.038}^*$ & $2.238_{\pm0.025}$ & $\mathbf{2.219}_{\pm0.023}^*$ & $1.524_{\pm0.022}$ & $\mathbf{1.510}_{\pm0.025}^*$ \\
      \midrule
      Bagging & $\mathbf{0.985}_{\pm0.010}$ & $0.986_{\pm0.009}$ & $\mathbf{0.800}_{\pm0.008}^*$ & $0.804_{\pm0.007}$ & $\mathbf{0.707}_{\pm0.010}^*$ & $0.713_{\pm0.011}$ \\
      \midrule
      LightGBM & $\mathbf{0.976}_{\pm0.011}^*$      &   $0.982_{\pm0.011}$    &   $\mathbf{0.786}_{\pm0.008}^*$    &   $0.794_{\pm0.008}$    &  $\mathbf{0.676}_{\pm0.010}^*$     & $0.695_{\pm0.012}$ \\
      \midrule
      Total UPDRS & \multicolumn{2}{c}{RMSE} & \multicolumn{2}{c}{MAE} & \multicolumn{2}{c}{MedianAE} \\
      \midrule
      SVR   & $2.581_{\pm0.130}$ & $\mathbf{2.323}_{\pm0.077}^*$ & $0.977_{\pm0.014}$ & $\mathbf{0.927}_{\pm0.010}^*$ & $0.727_{\pm0.015}$ & $\mathbf{0.684}_{\pm0.008}^*$ \\
      \midrule
      NN    & $0.988_{\pm0.003}$ & $\mathbf{0.981}_{\pm0.004}^*$ & $\mathbf{0.798}_{\pm0.002}^*$ & $0.804_{\pm0.004}$ & $\mathbf{0.701}_{\pm0.008}^*$ & $0.713_{\pm0.006}$ \\
      \midrule
      GPR   & $3.110_{\pm0.036}$ & $\mathbf{3.085}_{\pm0.032}^*$ & $2.187_{\pm0.023}$ & $\mathbf{2.171}_{\pm0.022}^*$ & $1.490_{\pm0.027}$ & $\mathbf{1.477}_{\pm0.024}^*$ \\
      \midrule
      Bagging & $1.019_{\pm0.010}$ & $\mathbf{1.016}_{\pm0.010}^*$ & $\mathbf{0.798}_{\pm0.009}^*$ & $0.800_{\pm0.007}$ & $\mathbf{0.652}_{\pm0.009}^*$ & $0.660_{\pm0.007}$ \\
      \midrule
      LightGBM &  $\mathbf{0.999}_{\pm0.009}$      &  $0.999_{\pm0.010}$      &  $\mathbf{0.784}_{\pm0.007}^*$      &   $0.788_{\pm0.007}$     &   $\mathbf{0.643}_{\pm0.013}^*$     &  $0.653_{\pm0.009}$ \\
      \bottomrule
      \end{tabular}%
    \label{tab:Binning Feature}%
  \end{table*}%

Some NoRo prediction errors of downstream models (NN, Bagging, LightGBM) are significantly higher than baseline. Thus, the feature augmentation method with other binning feature is less noise-robust than with the selected feature \{DFA\}, which proves the effectiveness of the binning feature selection module.

\subsection{Feature Space Observation (RQ5)} \label{sec:Feature Space}

\begin{figure*}[htbp]
    \centering
        \subfigure[Original Features]
        {\includegraphics[width=0.45\textwidth]{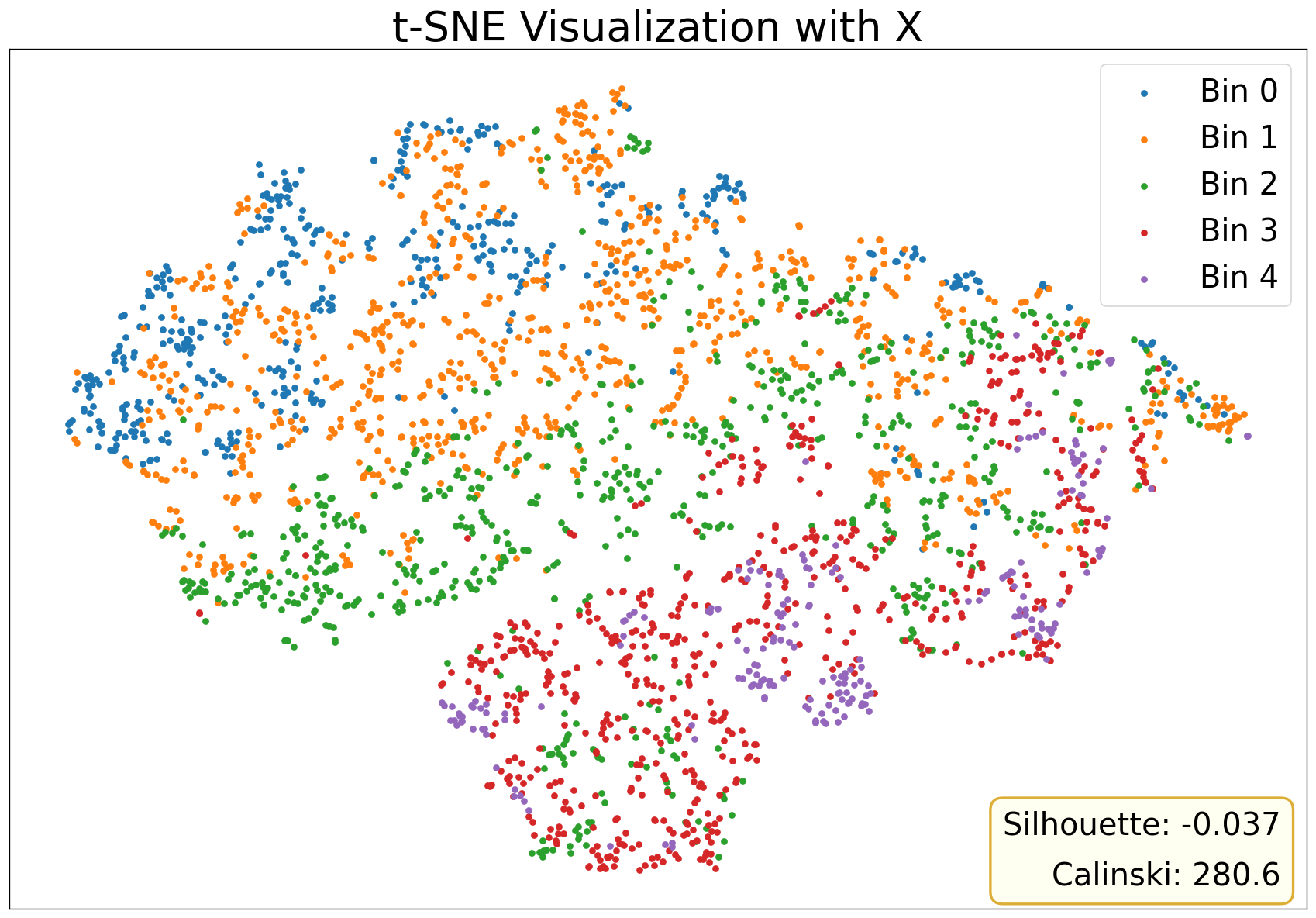}
        \label{fig:6.a}}
        \subfigure[Augmented Features]
        {\includegraphics[width=0.45\textwidth]{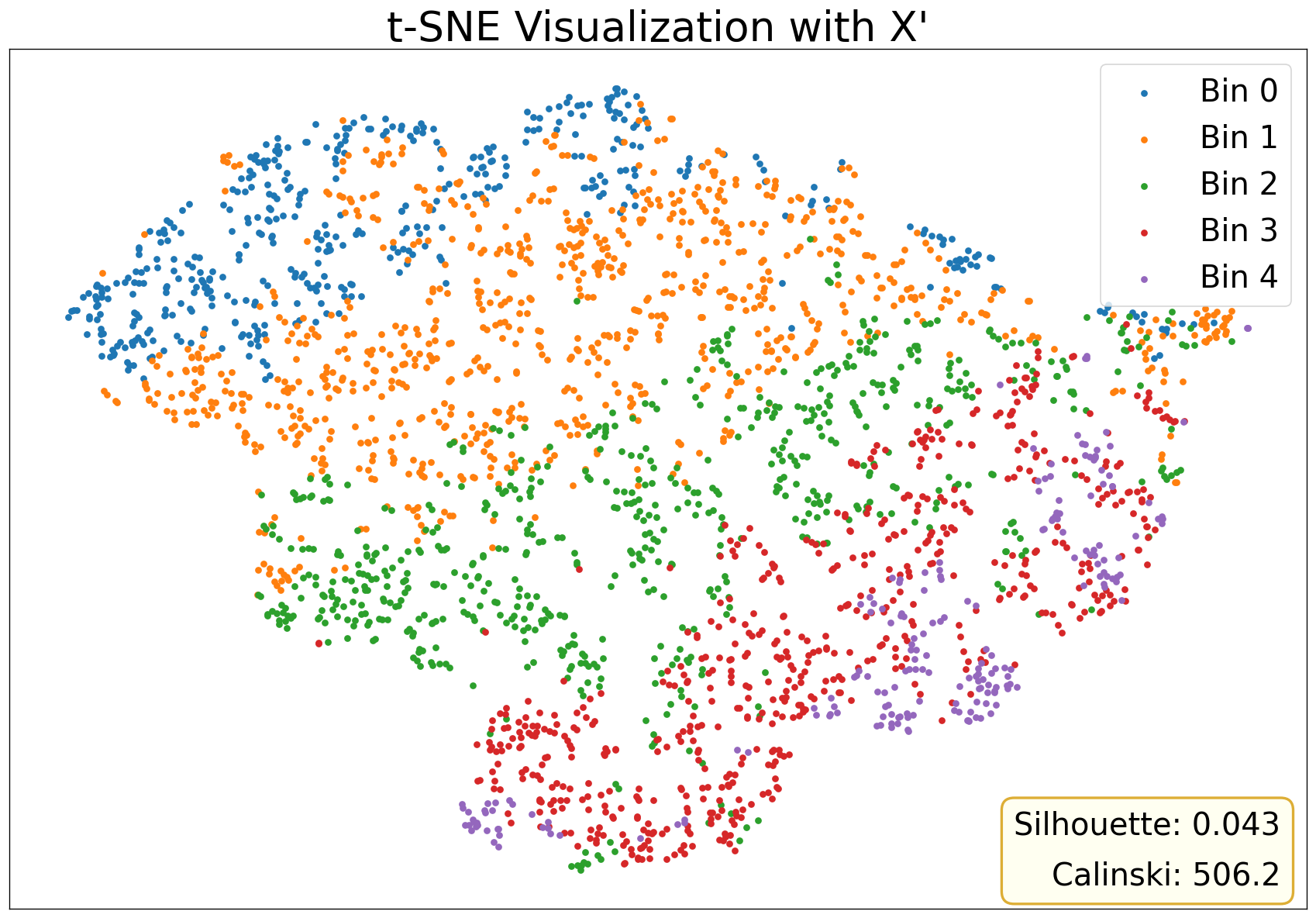}\label{fig:6.b}}
        
        \subfigure[Noisy Features]
        {\includegraphics[width=0.45\textwidth]{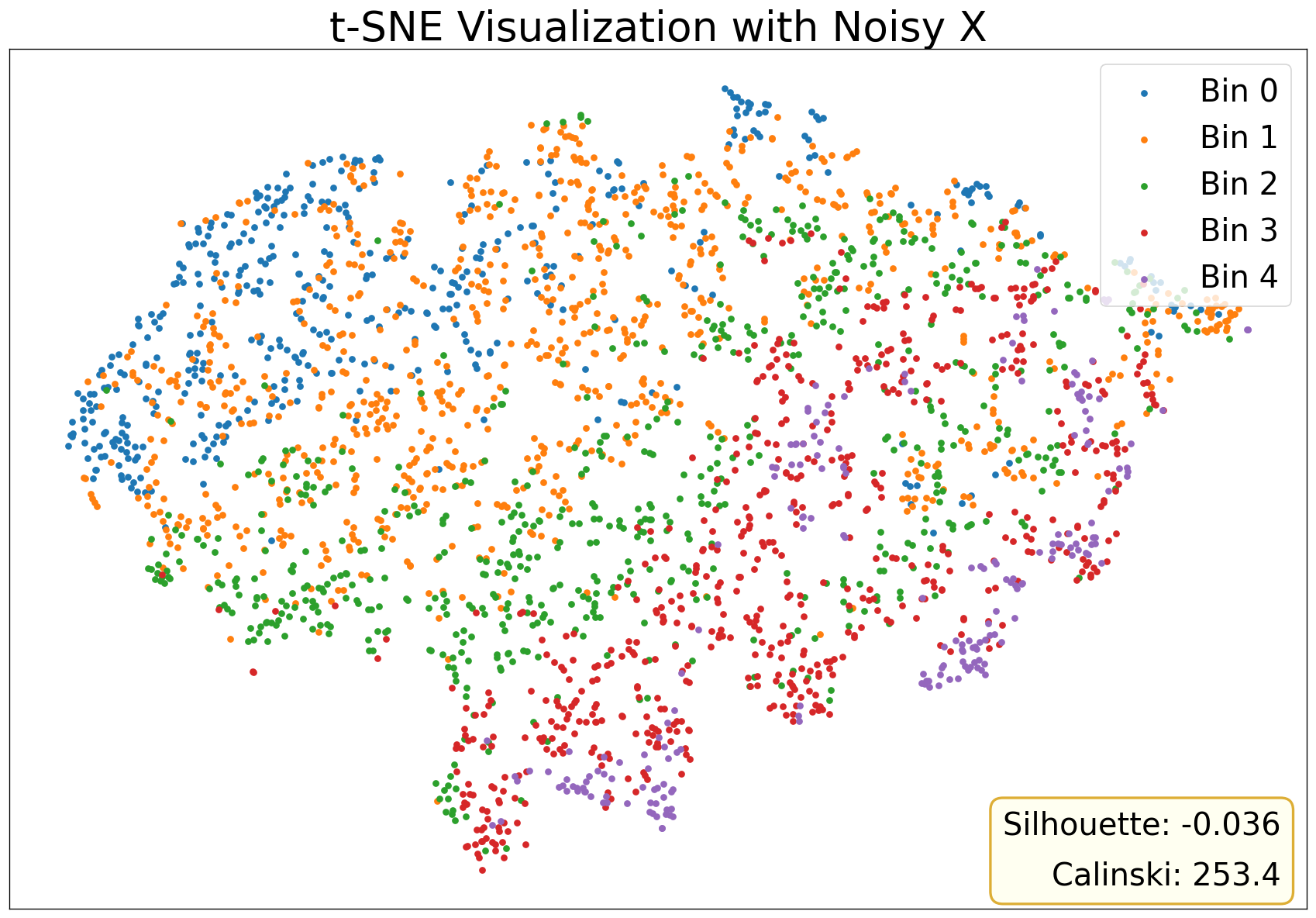}
        \label{fig:6.c}}
        \subfigure[Noisy Augmented Features]
        {\includegraphics[width=0.45\textwidth]{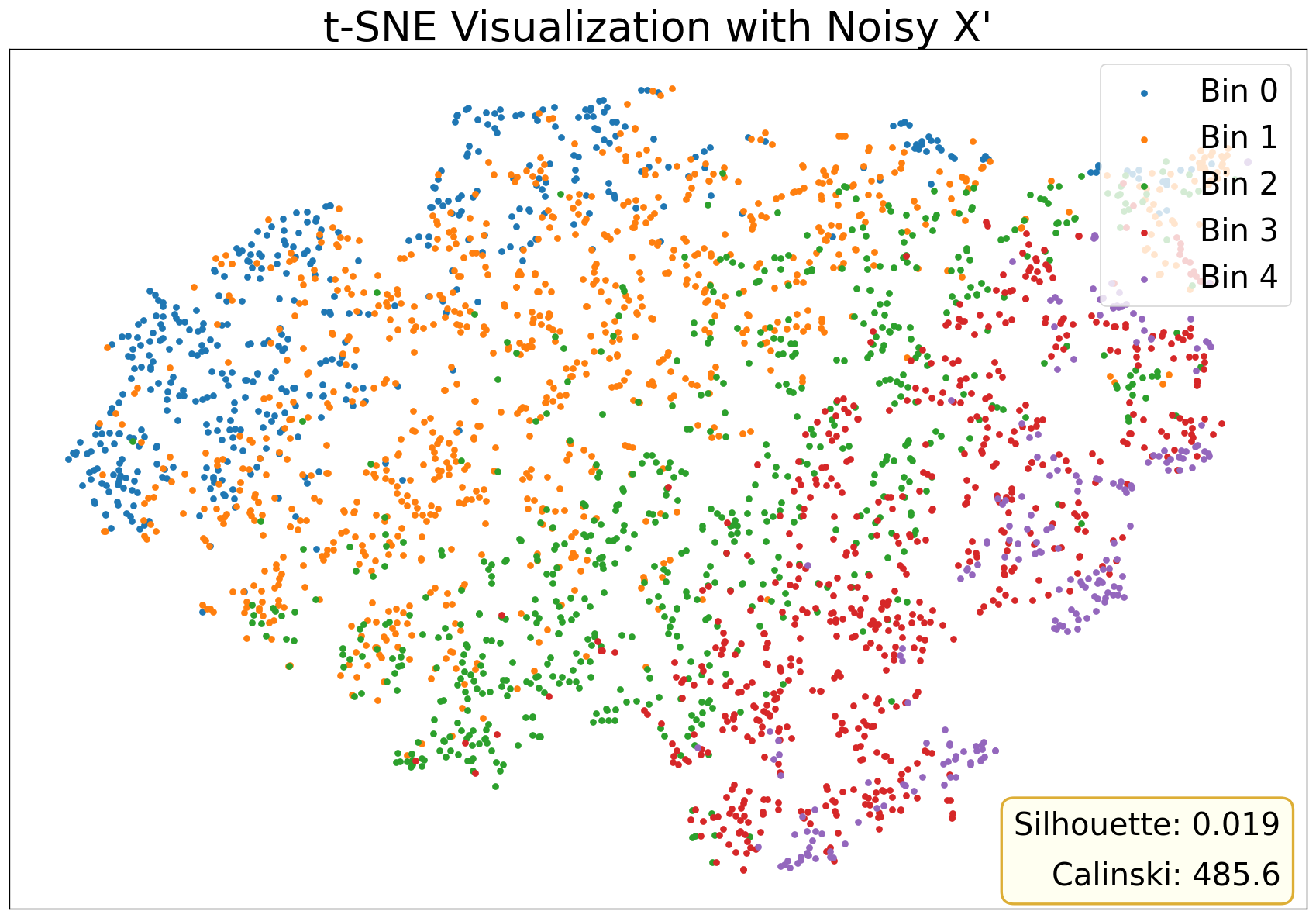}\label{fig:6.d}}
             
    \caption{T-SNE visualization of feature spaces. Points with the same color are the samples of the same bin. One color represents one certain bin. The relative positions between the points indicates the relative positions between them in original feature space or augmented feature space. Corresponding metrics are reported in each subplot, where higher Silhouette score and higher Calinski-Harabasz index indicate better results.}
    \label{fig:t-SNE}
\end{figure*}

To observe the augmented feature space, Fig. \ref{fig:t-SNE} presents the t-SNE visualizations \cite{van2008visualizing} of original and the augmented feature space for the test speech features without or with extra noise (SNR=30dB). Noisy speech features retain original feature bin labels, demonstrating noise impact through controlled label persistence. Three common-used metrics (i.e., Silhouette Score and Calinski-Harabasz Index) for unsupervised learning are calculated in Fig. \ref{fig:t-SNE} to evaluate the binning results quantitatively.

With noise introduction (from Fig.~\ref{fig:6.a} to \ref{fig:6.c}), Bin 3 shows distortion, while augmented features (from Fig.~\ref{fig:6.c} to \ref{fig:6.d}) reduce the distortion of Bin 3. Compare Fig. \ref{fig:6.a} and \ref{fig:6.c} with Fig. \ref{fig:6.b} and \ref{fig:6.d}, the binning results in augmented feature space are better than in original feature space, indicating lower distance between the same-bin samples, while higher distance between the cross-bin samples.

More specifically, compare Fig. \ref{fig:6.a} with Fig. \ref{fig:6.c}, after extra noise is introduced, Calinski-Harabasz index decreases by $27.2$. While compare Fig. \ref{fig:6.b} with Fig. \ref{fig:6.d}, Calinski-Harabasz index decreases by $19.6$, which is lower than the former decrease of $27.2$. This phenomenon indicates that NoRo successfully preserves the discriminative nature of the samples in the augmented feature space. But for Silhouette score, compare Fig. \ref{fig:6.a} with Fig. \ref{fig:6.c}, Silhouette score is almost the same, which may stem from the randomness of the extra random noise.

Thus, the augmented feature space is more noise-robust than the original feature space as expected, which explains the effectiveness of NoRo from the perspective of feature space.



\section{Conclusion}
In this work, we proposed a noise-robust UPDRS prediction framework (called NoRo), which achieved consistent noise robustness enhancements across various downstream prediction models, reducing prediction errors by up to more than 10\% to 40\%. Concretely, the idea of NoRo is to leverage contrastive learning and the continuous values of original features to construct contrastive pairs for training a set of noise-robust features in an unsupervised learning paradigm. These noise-robust features make the samples (i.e., PD patients) with similar features in the original feature space closer in the augmented feature space, and push samples away from each other if dissimilar, thereby increasing the discriminative nature of the samples in the augmented feature space even under some noisy environments. 

Comprehensive experiments, such as quantitative analysis, qualitative analysis, and visualization of feature spaces, were conducted and have demonstrated the effectiveness and robustness of the proposed NoRo framework. It is interesting to observe that, with or without NoRo, the ensemble models achieve better performance than the simple models. One future work is thus to integrate the ensemble mechanism into the proposed framework to investigate whether the ensemble mechanism can further boost the performance of RoNo. Besides, the existing methods for PD telemonitoring UPDRS prediction mainly consider the speech signal features, therefore another promising future work is to also include other useful features like age and gender.








\printcredits

\bibliographystyle{model1-num-names}

\bibliography{cas-refs}

\begin{thebibliography}{38}
\expandafter\ifx\csname natexlab\endcsname\relax\def\natexlab#1{#1}\fi
\providecommand{\url}[1]{\texttt{#1}}
\providecommand{\href}[2]{#2}
\providecommand{\path}[1]{#1}
\providecommand{\DOIprefix}{doi:}
\providecommand{\ArXivprefix}{arXiv:}
\providecommand{\URLprefix}{URL: }
\providecommand{\Pubmedprefix}{pmid:}
\providecommand{\doi}[1]{\href{http://dx.doi.org/#1}{\path{#1}}}
\providecommand{\Pubmed}[1]{\href{pmid:#1}{\path{#1}}}
\providecommand{\bibinfo}[2]{#2}
\ifx\xfnm\relax \def\xfnm[#1]{\unskip,\space#1}\fi
\bibitem[{Kalia and Lang(2015)}]{kalia2015parkinson}
\bibinfo{author}{L.~V. Kalia}, \bibinfo{author}{A.~E. Lang},
\newblock \bibinfo{title}{Parkinson's disease},
\newblock \bibinfo{journal}{The Lancet} \bibinfo{volume}{386} (\bibinfo{year}{2015}) \bibinfo{pages}{896--912}.
\bibitem[{Ascherio and Schwarzschild(2016)}]{ascherio2016epidemiology}
\bibinfo{author}{A.~Ascherio}, \bibinfo{author}{M.~A. Schwarzschild},
\newblock \bibinfo{title}{The epidemiology of parkinson's disease: risk factors and prevention},
\newblock \bibinfo{journal}{The Lancet Neurology} \bibinfo{volume}{15} (\bibinfo{year}{2016}) \bibinfo{pages}{1257--1272}.
\bibitem[{Bloem et~al.(2021)Bloem, Okun, and Klein}]{bloem2021parkinson}
\bibinfo{author}{B.~R. Bloem}, \bibinfo{author}{M.~S. Okun}, \bibinfo{author}{C.~Klein},
\newblock \bibinfo{title}{Parkinson's disease},
\newblock \bibinfo{journal}{The Lancet} \bibinfo{volume}{397} (\bibinfo{year}{2021}) \bibinfo{pages}{2284--2303}.
\bibitem[{Reeve et~al.(2014)Reeve, Simcox, and Turnbull}]{reeve2014ageing}
\bibinfo{author}{A.~Reeve}, \bibinfo{author}{E.~Simcox}, \bibinfo{author}{D.~Turnbull},
\newblock \bibinfo{title}{Ageing and parkinson's disease: why is advancing age the biggest risk factor?},
\newblock \bibinfo{journal}{Ageing research reviews} \bibinfo{volume}{14} (\bibinfo{year}{2014}) \bibinfo{pages}{19--30}.
\bibitem[{Samii et~al.(2004)Samii, Nutt, and Ransom}]{samii2004parkinson}
\bibinfo{author}{A.~Samii}, \bibinfo{author}{J.~G. Nutt}, \bibinfo{author}{B.~R. Ransom},
\newblock \bibinfo{title}{Parkinson's disease},
\newblock \bibinfo{journal}{The Lancet} \bibinfo{volume}{363} (\bibinfo{year}{2004}) \bibinfo{pages}{1783--1793}.
\bibitem[{AE(1998)}]{ae1998parkinson}
\bibinfo{author}{L.~AE},
\newblock \bibinfo{title}{Parkinson’s disease. first of two parts},
\newblock \bibinfo{journal}{N. Engl. J. Med.} \bibinfo{volume}{339} (\bibinfo{year}{1998}) \bibinfo{pages}{1044--1053}.
\bibitem[{Halliday and McCann(2010)}]{halliday2010progression}
\bibinfo{author}{G.~M. Halliday}, \bibinfo{author}{H.~McCann},
\newblock \bibinfo{title}{The progression of pathology in parkinson's disease},
\newblock \bibinfo{journal}{Annals of the New York Academy of Sciences} \bibinfo{volume}{1184} (\bibinfo{year}{2010}) \bibinfo{pages}{188--195}.
\bibitem[{Jankovic(2008)}]{jankovic2008parkinson}
\bibinfo{author}{J.~Jankovic},
\newblock \bibinfo{title}{Parkinson’s disease: clinical features and diagnosis},
\newblock \bibinfo{journal}{Journal of neurology, neurosurgery \& psychiatry} \bibinfo{volume}{79} (\bibinfo{year}{2008}) \bibinfo{pages}{368--376}.
\bibitem[{Poewe and Mahlknecht(2009)}]{poewe2009clinical}
\bibinfo{author}{W.~Poewe}, \bibinfo{author}{P.~Mahlknecht},
\newblock \bibinfo{title}{The clinical progression of parkinson's disease},
\newblock \bibinfo{journal}{Parkinsonism \& related disorders} \bibinfo{volume}{15} (\bibinfo{year}{2009}) \bibinfo{pages}{S28--S32}.
\bibitem[{Goetz et~al.(2009)Goetz, Stebbins, Wolff, DeLeeuw, Bronte-Stewart, Elble, Hallett, Nutt, Ramig, Sanger et~al.}]{goetz2009testing}
\bibinfo{author}{C.~G. Goetz}, \bibinfo{author}{G.~T. Stebbins}, \bibinfo{author}{D.~Wolff}, \bibinfo{author}{W.~DeLeeuw}, \bibinfo{author}{H.~Bronte-Stewart}, \bibinfo{author}{R.~Elble}, \bibinfo{author}{M.~Hallett}, \bibinfo{author}{J.~Nutt}, \bibinfo{author}{L.~Ramig}, \bibinfo{author}{T.~Sanger}, et~al.,
\newblock \bibinfo{title}{Testing objective measures of motor impairment in early parkinson's disease: Feasibility study of an at-home testing device},
\newblock \bibinfo{journal}{Movement Disorders} \bibinfo{volume}{24} (\bibinfo{year}{2009}) \bibinfo{pages}{551--556}.
\bibitem[{Tsanas et~al.(2009)Tsanas, Little, McSharry, and Ramig}]{tsanas2009accurate}
\bibinfo{author}{A.~Tsanas}, \bibinfo{author}{M.~Little}, \bibinfo{author}{P.~McSharry}, \bibinfo{author}{L.~Ramig},
\newblock \bibinfo{title}{Accurate telemonitoring of parkinson’s disease progression by non-invasive speech tests},
\newblock \bibinfo{journal}{Nature Precedings}  (\bibinfo{year}{2009}) \bibinfo{pages}{1--1}.
\bibitem[{on~Rating Scales~for Parkinson's~Disease(2003)}]{movement2003unified}
\bibinfo{author}{M.~D. S. T.~F. on~Rating Scales~for Parkinson's~Disease},
\newblock \bibinfo{title}{The unified parkinson's disease rating scale (updrs): status and recommendations},
\newblock \bibinfo{journal}{Movement Disorders} \bibinfo{volume}{18} (\bibinfo{year}{2003}) \bibinfo{pages}{738--750}.
\bibitem[{Nilashi et~al.(2018)Nilashi, Ibrahim, Ahmadi, Shahmoradi, and Farahmand}]{nilashi2018hybrid}
\bibinfo{author}{M.~Nilashi}, \bibinfo{author}{O.~Ibrahim}, \bibinfo{author}{H.~Ahmadi}, \bibinfo{author}{L.~Shahmoradi}, \bibinfo{author}{M.~Farahmand},
\newblock \bibinfo{title}{A hybrid intelligent system for the prediction of parkinson's disease progression using machine learning techniques},
\newblock \bibinfo{journal}{Biocybernetics and Biomedical Engineering} \bibinfo{volume}{38} (\bibinfo{year}{2018}) \bibinfo{pages}{1--15}.
\bibitem[{Nilashi et~al.(2019)Nilashi, Ibrahim, Samad, Ahmadi, Shahmoradi, and Akbari}]{nilashi2019analytical}
\bibinfo{author}{M.~Nilashi}, \bibinfo{author}{O.~Ibrahim}, \bibinfo{author}{S.~Samad}, \bibinfo{author}{H.~Ahmadi}, \bibinfo{author}{L.~Shahmoradi}, \bibinfo{author}{E.~Akbari},
\newblock \bibinfo{title}{An analytical method for measuring the parkinson’s disease progression: A case on a parkinson’s telemonitoring dataset},
\newblock \bibinfo{journal}{Measurement} \bibinfo{volume}{136} (\bibinfo{year}{2019}) \bibinfo{pages}{545--557}.
\bibitem[{Nilashi et~al.(2023)Nilashi, Abumalloh, Alyami, Alghamdi, and Alrizq}]{nilashi2023parkinson}
\bibinfo{author}{M.~Nilashi}, \bibinfo{author}{R.~A. Abumalloh}, \bibinfo{author}{S.~Alyami}, \bibinfo{author}{A.~Alghamdi}, \bibinfo{author}{M.~Alrizq},
\newblock \bibinfo{title}{Parkinson’s disease diagnosis using laplacian score, gaussian process regression and self-organizing maps},
\newblock \bibinfo{journal}{Brain Sciences} \bibinfo{volume}{13} (\bibinfo{year}{2023}) \bibinfo{pages}{543}.
\bibitem[{Zogaan et~al.(2024)Zogaan, Nilashi, Ahmadi, Abumalloh, Alrizq, Abosaq, and Alghamdi}]{zogaan2024combined}
\bibinfo{author}{W.~A. Zogaan}, \bibinfo{author}{M.~Nilashi}, \bibinfo{author}{H.~Ahmadi}, \bibinfo{author}{R.~A. Abumalloh}, \bibinfo{author}{M.~Alrizq}, \bibinfo{author}{H.~Abosaq}, \bibinfo{author}{A.~Alghamdi},
\newblock \bibinfo{title}{A combined method of optimized learning vector quantization and neuro-fuzzy techniques for predicting unified parkinson's disease rating scale using vocal features},
\newblock \bibinfo{journal}{MethodsX} \bibinfo{volume}{12} (\bibinfo{year}{2024}) \bibinfo{pages}{102553}.
\bibitem[{Vats et~al.(2023)Vats, Blouria, and Sasikala}]{vats2023predicting}
\bibinfo{author}{A.~Vats}, \bibinfo{author}{A.~Blouria}, \bibinfo{author}{R.~Sasikala},
\newblock \bibinfo{title}{Predicting severity levels of parkinson’s disease from telemonitoring voice data},
\newblock in: \bibinfo{booktitle}{Inventive Systems and Control: Proceedings of ICISC 2023}, \bibinfo{publisher}{Springer}, \bibinfo{year}{2023}, pp. \bibinfo{pages}{839--853}.
\bibitem[{Nicastri et~al.(2004)Nicastri, Chiarella, Gallo, Catalano, Cassandro et~al.}]{nicastri2004multidimensional}
\bibinfo{author}{M.~Nicastri}, \bibinfo{author}{G.~Chiarella}, \bibinfo{author}{L.~Gallo}, \bibinfo{author}{M.~Catalano}, \bibinfo{author}{E.~Cassandro}, et~al.,
\newblock \bibinfo{title}{Multidimensional voice program (mdvp) and amplitude variation parameters in euphonic adult subjects. normative study},
\newblock \bibinfo{journal}{Acta Otorhinolaryngol Ital} \bibinfo{volume}{24} (\bibinfo{year}{2004}) \bibinfo{pages}{337--341}.
\bibitem[{Tsanas(2012)}]{tsanas2012accurate}
\bibinfo{author}{A.~Tsanas}, \bibinfo{title}{Accurate telemonitoring of Parkinson’s disease symptom severity using nonlinear speech signal processing and statistical machine learning}, Ph.D. thesis, Oxford University, UK, \bibinfo{year}{2012}.
\bibitem[{Zhao et~al.(2015)Zhao, Li, Li, and Wang}]{zhao2015householder}
\bibinfo{author}{Y.-P. Zhao}, \bibinfo{author}{B.~Li}, \bibinfo{author}{Y.-B. Li}, \bibinfo{author}{K.-K. Wang},
\newblock \bibinfo{title}{Householder transformation based sparse least squares support vector regression},
\newblock \bibinfo{journal}{Neurocomputing} \bibinfo{volume}{161} (\bibinfo{year}{2015}) \bibinfo{pages}{243--253}.
\bibitem[{Yoon and Li(2018)}]{yoon2018novel}
\bibinfo{author}{H.~Yoon}, \bibinfo{author}{J.~Li},
\newblock \bibinfo{title}{A novel positive transfer learning approach for telemonitoring of parkinson’s disease},
\newblock \bibinfo{journal}{IEEE Transactions on Automation Science and Engineering} \bibinfo{volume}{16} (\bibinfo{year}{2018}) \bibinfo{pages}{180--191}.
\bibitem[{Nilashi et~al.(2023)Nilashi, Abumalloh, Yusuf, Thi, Alsulami, Abosaq, Alyami, and Alghamdi}]{nilashi2023early}
\bibinfo{author}{M.~Nilashi}, \bibinfo{author}{R.~A. Abumalloh}, \bibinfo{author}{S.~Y.~M. Yusuf}, \bibinfo{author}{H.~H. Thi}, \bibinfo{author}{M.~Alsulami}, \bibinfo{author}{H.~Abosaq}, \bibinfo{author}{S.~Alyami}, \bibinfo{author}{A.~Alghamdi},
\newblock \bibinfo{title}{Early diagnosis of parkinson’s disease: A combined method using deep learning and neuro-fuzzy techniques},
\newblock \bibinfo{journal}{Computational biology and chemistry} \bibinfo{volume}{102} (\bibinfo{year}{2023}) \bibinfo{pages}{107788}.
\bibitem[{Shorten and Khoshgoftaar(2019)}]{shorten2019survey}
\bibinfo{author}{C.~Shorten}, \bibinfo{author}{T.~M. Khoshgoftaar},
\newblock \bibinfo{title}{A survey on image data augmentation for deep learning},
\newblock \bibinfo{journal}{Journal of big data} \bibinfo{volume}{6} (\bibinfo{year}{2019}) \bibinfo{pages}{1--48}.
\bibitem[{Ko et~al.(2015)Ko, Peddinti, Povey, and Khudanpur}]{ko2015audio}
\bibinfo{author}{T.~Ko}, \bibinfo{author}{V.~Peddinti}, \bibinfo{author}{D.~Povey}, \bibinfo{author}{S.~Khudanpur},
\newblock \bibinfo{title}{Audio augmentation for speech recognition.},
\newblock in: \bibinfo{booktitle}{Interspeech}, volume \bibinfo{volume}{2015}, \bibinfo{year}{2015}, p. \bibinfo{pages}{3586}.
\bibitem[{Xiao et~al.(2020)Xiao, Wang, Efros, and Darrell}]{xiao2020should}
\bibinfo{author}{T.~Xiao}, \bibinfo{author}{X.~Wang}, \bibinfo{author}{A.~A. Efros}, \bibinfo{author}{T.~Darrell},
\newblock \bibinfo{title}{What should not be contrastive in contrastive learning},
\newblock \bibinfo{journal}{arXiv preprint arXiv:2008.05659}  (\bibinfo{year}{2020}).
\bibitem[{Zhang et~al.(2023)Zhang, Zhu, Song, Koniusz, and King}]{zhang2023spectral}
\bibinfo{author}{Y.~Zhang}, \bibinfo{author}{H.~Zhu}, \bibinfo{author}{Z.~Song}, \bibinfo{author}{P.~Koniusz}, \bibinfo{author}{I.~King},
\newblock \bibinfo{title}{Spectral feature augmentation for graph contrastive learning and beyond},
\newblock in: \bibinfo{booktitle}{Proceedings of the AAAI Conference on Artificial Intelligence}, volume~\bibinfo{volume}{37}, \bibinfo{year}{2023}, pp. \bibinfo{pages}{11289--11297}.
\bibitem[{Zhu et~al.(2021)Zhu, Xu, Yu, Liu, Wu, and Wang}]{zhu2021graph}
\bibinfo{author}{Y.~Zhu}, \bibinfo{author}{Y.~Xu}, \bibinfo{author}{F.~Yu}, \bibinfo{author}{Q.~Liu}, \bibinfo{author}{S.~Wu}, \bibinfo{author}{L.~Wang},
\newblock \bibinfo{title}{Graph contrastive learning with adaptive augmentation},
\newblock in: \bibinfo{booktitle}{Proceedings of the web conference 2021}, \bibinfo{year}{2021}, pp. \bibinfo{pages}{2069--2080}.
\bibitem[{Zhang et~al.(2024)Zhang, Hou, Jiang, Zhang, Zhou, Tang, and Lv}]{zhang2024label}
\bibinfo{author}{T.~Zhang}, \bibinfo{author}{C.~Hou}, \bibinfo{author}{R.~Jiang}, \bibinfo{author}{X.~Zhang}, \bibinfo{author}{C.~Zhou}, \bibinfo{author}{K.~Tang}, \bibinfo{author}{H.~Lv},
\newblock \bibinfo{title}{Label informed contrastive pretraining for node importance estimation on knowledge graphs},
\newblock \bibinfo{journal}{IEEE Transactions on Neural Networks and Learning Systems}  (\bibinfo{year}{2024}).
\bibitem[{Wang et~al.(2023)Wang, Zhou, Shen, and Jia}]{wang2023analysis}
\bibinfo{author}{S.~Wang}, \bibinfo{author}{T.~Zhou}, \bibinfo{author}{Z.~Shen}, \bibinfo{author}{Z.~Jia},
\newblock \bibinfo{title}{Analysis of augmentations in contrastive learning for parkinson’s disease diagnosis},
\newblock in: \bibinfo{booktitle}{International Conference on Artificial Neural Networks}, \bibinfo{organization}{Springer}, \bibinfo{year}{2023}, pp. \bibinfo{pages}{37--50}.
\bibitem[{Hasan et~al.(2016)Hasan, Nasser, Ahmad, and Molla}]{hasan2016feature}
\bibinfo{author}{M.~A.~M. Hasan}, \bibinfo{author}{M.~Nasser}, \bibinfo{author}{S.~Ahmad}, \bibinfo{author}{K.~I. Molla},
\newblock \bibinfo{title}{Feature selection for intrusion detection using random forest},
\newblock \bibinfo{journal}{Journal of information security} \bibinfo{volume}{7} (\bibinfo{year}{2016}) \bibinfo{pages}{129--140}.
\bibitem[{Tsanas and Little(2009)}]{misc_parkinsons_telemonitoring_189}
\bibinfo{author}{A.~Tsanas}, \bibinfo{author}{M.~Little}, \bibinfo{title}{{Parkinsons Telemonitoring}}, \bibinfo{howpublished}{UCI Machine Learning Repository}, \bibinfo{year}{2009}. \bibinfo{note}{{DOI}: https://doi.org/10.24432/C5ZS3N}.
\bibitem[{Mivule(2013)}]{mivule2013utilizing}
\bibinfo{author}{K.~Mivule},
\newblock \bibinfo{title}{Utilizing noise addition for data privacy, an overview},
\newblock \bibinfo{journal}{arXiv preprint arXiv:1309.3958}  (\bibinfo{year}{2013}).
\bibitem[{Hemmerling et~al.(2023)Hemmerling, W{\'o}jcik-Pedziwiatr, Jaci{\'o}w, Zi{\'o}{\l}ko, and Igras-Cybulska}]{hemmerling2023monitoring}
\bibinfo{author}{D.~Hemmerling}, \bibinfo{author}{M.~W{\'o}jcik-Pedziwiatr}, \bibinfo{author}{P.~Jaci{\'o}w}, \bibinfo{author}{B.~Zi{\'o}{\l}ko}, \bibinfo{author}{M.~Igras-Cybulska},
\newblock \bibinfo{title}{Monitoring of parkinson’s disease progression based on speech signal},
\newblock in: \bibinfo{booktitle}{2023 6th International Conference on Information and Computer Technologies (ICICT)}, \bibinfo{organization}{IEEE}, \bibinfo{year}{2023}, pp. \bibinfo{pages}{132--137}.
\bibitem[{Breiman(1996)}]{breiman1996bagging}
\bibinfo{author}{L.~Breiman},
\newblock \bibinfo{title}{Bagging predictors},
\newblock \bibinfo{journal}{Machine learning} \bibinfo{volume}{24} (\bibinfo{year}{1996}) \bibinfo{pages}{123--140}.
\bibitem[{Ke et~al.(2017)Ke, Meng, Finley, Wang, Chen, Ma, Ye, and Liu}]{ke2017lightgbm}
\bibinfo{author}{G.~Ke}, \bibinfo{author}{Q.~Meng}, \bibinfo{author}{T.~Finley}, \bibinfo{author}{T.~Wang}, \bibinfo{author}{W.~Chen}, \bibinfo{author}{W.~Ma}, \bibinfo{author}{Q.~Ye}, \bibinfo{author}{T.-Y. Liu},
\newblock \bibinfo{title}{Lightgbm: A highly efficient gradient boosting decision tree},
\newblock \bibinfo{journal}{Advances in neural information processing systems} \bibinfo{volume}{30} (\bibinfo{year}{2017}).
\bibitem[{Nilashi et~al.(2022)Nilashi, Abumalloh, Minaei-Bidgoli, Samad, Yousoof~Ismail, Alhargan, and Abdu~Zogaan}]{nilashi2022predicting}
\bibinfo{author}{M.~Nilashi}, \bibinfo{author}{R.~A. Abumalloh}, \bibinfo{author}{B.~Minaei-Bidgoli}, \bibinfo{author}{S.~Samad}, \bibinfo{author}{M.~Yousoof~Ismail}, \bibinfo{author}{A.~Alhargan}, \bibinfo{author}{W.~Abdu~Zogaan},
\newblock \bibinfo{title}{Predicting parkinson’s disease progression: Evaluation of ensemble methods in machine learning},
\newblock \bibinfo{journal}{Journal of healthcare engineering} \bibinfo{volume}{2022} (\bibinfo{year}{2022}) \bibinfo{pages}{2793361}.
\bibitem[{Ku et~al.(1966)}]{ku1966notes}
\bibinfo{author}{H.~H. Ku}, et~al.,
\newblock \bibinfo{title}{Notes on the use of propagation of error formulas},
\newblock \bibinfo{journal}{Journal of Research of the National Bureau of Standards} \bibinfo{volume}{70} (\bibinfo{year}{1966}).
\bibitem[{Van~der Maaten and Hinton(2008)}]{van2008visualizing}
\bibinfo{author}{L.~Van~der Maaten}, \bibinfo{author}{G.~Hinton},
\newblock \bibinfo{title}{Visualizing data using t-sne.},
\newblock \bibinfo{journal}{Journal of machine learning research} \bibinfo{volume}{9} (\bibinfo{year}{2008}).

\end{thebibliography}

\appendix

\section{Appendix}

\subsection{Feature Selection Results}\label{sec:RF results}

\begin{figure}[htbp]
    \centering

    \subfigure[Motor UPDRS]
    	{\includegraphics[width=0.45\textwidth]{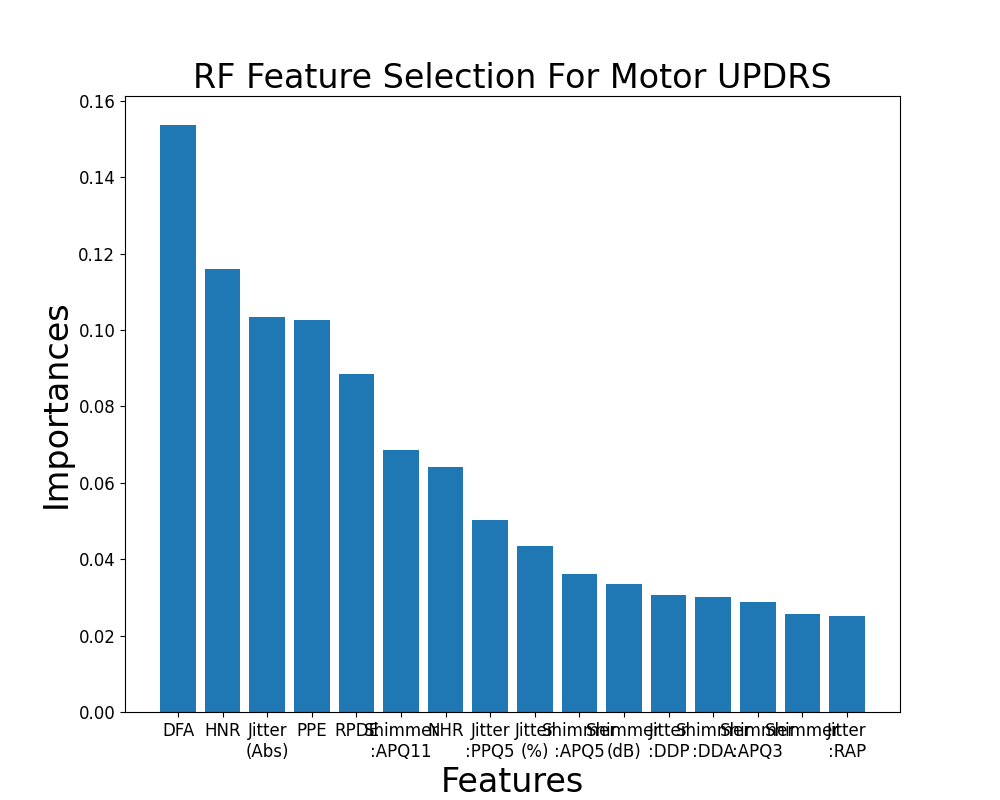}}

    \subfigure[Total UPDRS]
    	{\includegraphics[width=0.45\textwidth]{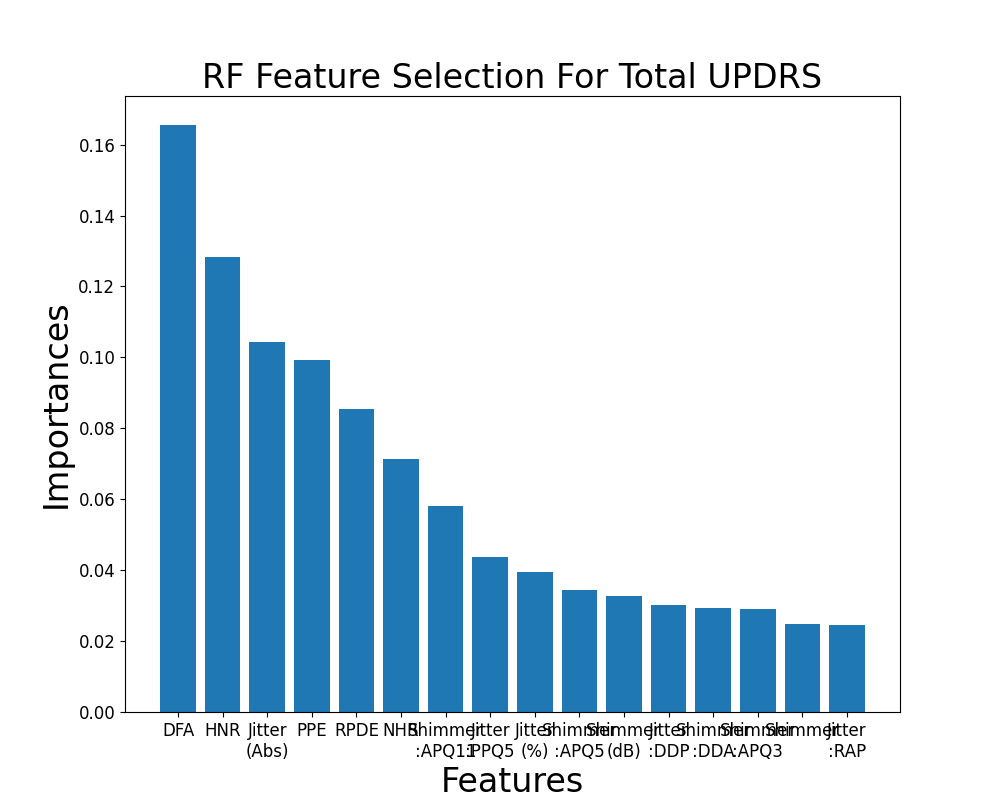}}

    \caption{Importance Scores by Random Forest}
    \label{fig:random forest}
\end{figure}

As mentioned in Section \ref{sec:feature selection}, the MDI importance scores of each feature dimension calculated by the random forest algorithm for both Motor and Total UPDRS are shown in Fig. \ref{fig:random forest}. The importance scores come from the average importance scores of 10 repeated trials with different random seeds.

Therefore, the most important feature dimension, namely \{DFA\}, is selected.









\subsection{Hyperparameter Settings}

\subsubsection{MLP Training Process}\label{app:MLP Training}

Adam optimizer is employed with an initial learning rate of $1\times {10}^{-3}$, and the loss function used is the CL Loss defined in Eq. \ref{eq:CLLoss}. The training hyperparameters are detailed in Tab. \ref{tab:Hyperparameter_Setting}.

\begin{table}[htbp]
    \centering
    \caption{Hyperparameter Setting}
    \begin{tabular}{c|c}
    \toprule
       Hyperparameter  &  Setting\\
    \midrule
        Optimizer & Adam \\
       Initial Learning Rate  &  $1\times {10}^{-3}$\\
       Activation Fuction & Tanh \\
       batch size & 2700 (Whole Training Set) \\
       epochs & 2000 \\
       gradient\_clip & 1.0 \\
       random seed & 2024 \\
       $K$ & 5 (Adjustable) \\
    \bottomrule
    \end{tabular}
    \label{tab:Hyperparameter_Setting}
\end{table}

As mentioned in Section \ref{section:Dataset}, a 10-fold cross-validation approach is employed. For each split of the training and validation sets, the model is trained for 200 epochs, adding to 2000 epochs in total. To prevent gradient explosion, gradient clipping is applied.

\subsubsection{Downstream Models}\label{app:downstream}

The hyperparameters of downstream models are detailed in Tab. \ref{tab:Downstream_Hyperparameter_Setting}.

\begin{table}[htbp]
    \centering
    \caption{Downstream models Hyperparameter Setting}
      \begin{tabular}{c|c}
        \toprule
        Model  &  Setting\\
        \midrule
      SVR   & kernel='poly' \\
      \hline
      GaussianProcessRegressor & - \\
      \hline
      MLPRegressor(NN) & solver="sgd" \\
            & alpha=1e-3 \\
            & activation="relu" \\
            & hidden\_layer\_sizes=(32) \\
            & max\_iter=2000 \\
            & tol=1e-3 \\
            & random\_state=2024 \\
        \hline
      BaggingRegressor & random\_state=2024 \\
      \hline
      LightGBM & num\_leaves=31\\
                &learning\_rate=0.1 \\
                &random\_state=2024 \\   
        \hline
        ANFIS Ensemble & SVD\_dim=4\\     
      \bottomrule
      \end{tabular}%
    \label{tab:Downstream_Hyperparameter_Setting}%
  \end{table}%

\subsection{Experimental Environment}

The experiments can be conducted on both Windows and Linux operating systems. The training algorithm of the MLP projection encoder and ANFIS is developed based on Pytorch. Other downstream models are implemented through machine learning libraries such as scikit-learn or lightgbm. The environment configuration is shown in Tab. \ref{tab:Environment Configuration}.

\begin{table}[!htbp]
    \centering
    \caption{Environment Configuration}
    \renewcommand\tabcolsep{3.8pt}
    \renewcommand{\arraystretch}{1.5}
       { \begin{tabular}{c|cc}
            \toprule
               Name & Version & Build\\
               \midrule
                python & 3.11.0 & h7a1cb2a\_3\\
                torch & 2.5.1 & pypi\_0 \\
                cuda & 12.4 & 0\\
                scikit-learn & 1.5.2 & pypi\_0 \\
                lightgbm & 4.5.0 & pypi\_0 \\
                \bottomrule
            \end{tabular}}
            \label{tab:Environment Configuration}
\end{table}

\end{document}